\documentclass[lettersize,journal]{IEEEtran}
\usepackage{amsmath,amsfonts}
\usepackage{algorithmic}
\usepackage{algorithm}
\usepackage{array}
\usepackage[caption=false,font=normalsize,labelfont=sf,textfont=sf]{subfig}
\usepackage{textcomp}
\usepackage{stfloats}
\usepackage{url}
\usepackage{verbatim}
\usepackage{graphicx}
\usepackage{cite}
\usepackage{listings}
\usepackage{booktabs}
\usepackage{multirow}
\usepackage{xcolor}
\usepackage[table]{xcolor}
\usepackage{array}
\usepackage{orcidlink}
\usepackage{kotex}
\usepackage[most]{tcolorbox}

\begin{document}

\title{A Multifaceted Analysis of Negative Bias in \\ Large Language Models through the Lens of \\ Parametric Knowledge}

\author{Jongyoon Song~\orcidlink{0009-0004-0451-1913}, Sangwon Yu~\orcidlink{0009-0008-4430-2111}, 
and Sungroh Yoon~\orcidlink{0000-0002-2367-197X}%
\thanks{J. Song was with the Department of Electrical and Computer Engineering at Seoul National University, South Korea (\href{mailto:coms1580@gmail.com}{coms1580@gmail.com}).}
\thanks{S. Yu is with the Department of Electrical and Computer Engineering at Seoul National University, South Korea (\href{mailto:dbtkddnjs96@snu.ac.kr}{dbtkddnjs96@snu.ac.kr}).}
\thanks{
S. Yoon is with the Department of Electrical and Computer Engineering,  Interdisciplinary Program in Artificial Intelligence, ASRI, INMC, ISRC, and Institute of Engineering Research, Seoul National University, South Korea (\href{mailto:sryoon@snu.ac.kr}{sryoon@snu.ac.kr}).}
\thanks{
S. Yoon is the corresponding author.
}
}

\markboth{A Multifaceted Analysis of Negative Bias in Large Language Models through the Lens of Parametric Knowledge}{A Multifaceted Analysis of Negative Bias in Large Language Models through the Lens of Parametric Knowledge}

\maketitle

\begin{abstract}
Negative bias refers to the tendency of large language models (LLMs) to excessively generate negative responses in binary decision tasks (e.g., yes-no question answering). 
Previous research has focused on detecting and addressing negative attention heads that induce negative bias.
However, the underlying detailed factors influencing negative bias remain underexplored.
In this paper, we demonstrate that LLMs exhibit \textit{format-level negative bias}, meaning the prompt format more influences their responses than the semantics of the negative response.
For the fine-grained study of the negative bias, we introduce a pipeline for constructing the evaluation set, which systematically categorizes the dataset into three subsets based on the model's parametric knowledge: correct, incorrect, and insufficient relevant knowledge.
Through analysis of this evaluation set, we identify a shortcut behavior in which models tend to generate negative responses when they lack sufficient knowledge to answer a yes-no question, leading to negative bias.
We further examine how negative bias changes under various prompting scenarios related to parametric knowledge.  
We observe that providing relevant context and offering an ``I don't know'' option generally reduces negative bias, whereas chain-of-thought prompting tends to amplify the bias.  
Finally, we demonstrate that the degree of negative bias can vary depending on the type of prompt, which influences the direction of the response.
Our work reveals the various factors that influence negative bias, providing critical insights for mitigating it in LLMs.
\end{abstract}


\section{Introduction}
\IEEEPARstart{R}{ecent} advances in the capabilities and emergent abilities of large language models (LLMs) have led to rapid improvements in the performance of a wide range of natural language processing (NLP) tasks \cite{dubey2024llama,jiang2023mistral,yang2024qwen2,ouyang2022training,achiam2023gpt}. 
Leveraging their ability to follow instructions, LLMs are able to perform complex, previously unseen tasks, enabling human-like interactions \cite{brown2020language,wei2022finetuned,kojima2022large,madaan2024self}.

Despite these breakthroughs, LLMs still exhibit vulnerabilities in terms of reliability and safety \cite{kaufmann2023survey}. 
One critical issue is the hallucination problem, where the model generates content that contains misleading information, which does not correspond to the given context or real-world knowledge \cite{zhang2023siren}. 
Although the factors contributing to hallucinations in LLMs are complex and can vary depending on the characteristics of the task, which makes it an ongoing area of active research \cite{addlesee2024grounding,xie2024adaptive,su2024textttconflictbank,chuang2024doladecodingcontrastinglayers}, the hallucination problem is closely related to \textit{parametric knowledge}, the embedded knowledge within the model parameters. 
Recent studies have reported that when parametric knowledge contradicts real-world information or the input context, knowledge conflict arises, often leading to the generation of hallucinated content \cite{zhang2023siren, xie2024adaptive,cheang2023lms,xu2024knowledge,wang2024resolving}. 

\IEEEpubidadjcol In this paper, we focus on the hallucination problem that arises in binary decision tasks.
The binary decision task, which determines whether a given question is true or false, is a crucial component of interaction between users and LLMs.
This task encompasses yes-no question answering (QA), where the model responds with either \textit{Yes} or \textit{No} to determine the answer to a given yes-no question and answer verification query, which evaluates whether the prediction to a general question is correct.
Previous researches report that LLMs exhibit a \textit{negative bias} issue in binary decision tasks requiring complex reasoning, where models tend to return negative responses rather than positive responses \cite{song2024large,Song2024ICiD,yu2024correcting}.
This phenomenon contributes to hallucinations during binary decision tasks because it degrades the reliability of negative responses as the number of false negative predictions increases.

Although existing works addressing the negative bias problem propose effective frameworks to formulate and mitigate the issue, several important areas remain underexplored and warrant further investigation.
First, it remains unclear which specific behaviors of LLMs give rise to negative bias. For instance, there is a lack of detailed analysis distinguishing whether negative responses stem from genuinely pessimistic reasoning processes or simply from a preference for negative linguistic formats.
Second, existing studies provide limited insight into how negative bias manifests in relation to the presence or absence of parametric knowledge within the model. Given that errors in parametric knowledge constitute a significant cause of the hallucination problem, it is crucial to thoroughly investigate the relationship between negative bias and the model’s internal knowledge.
Lastly, current studies are constrained to a single prompting setting, leaving the influence of different prompting scenarios on negative bias largely unexplored.

We begin our study with the rationale that negative bias emerges when the model is prompted to generate responses including a negative format, such as \textit{No}. 
In other words, we demonstrate that LLMs exhibit format-level negative bias, focusing more on whether the format of the response is negative rather than whether the contextual meaning of the response is negative.
For example, LLMs may respond with ``No'' to the question, ``Say yes or no. Is 1+1 equal to 2?'', not because they really think that 1+1 equals something other than 2, but because they just prefer to generate the response ``No''.
Based on this statement, we define negative bias as the difference in a model’s preference for negative responses when answering the same question presented in two formats: a direct negative format (say yes or no) and an indirect negative format (say A or B when A is yes and B is no).

The main goal of this paper is to explore and identify the factors that contribute to negative bias stemming from the parametric knowledge of LLMs.
Most QA datasets do not contain annotations reflecting the parametric knowledge state of LLMs and are not originally designed as binary decision tasks.  
To address these limitations, we develop a pipeline that precisely partitions yes-no QA and short-answer QA datasets based on the model’s parametric knowledge state, and converts them into a binary decision task format.  
Specifically, we probe the parametric knowledge of LLMs while minimizing sources of bias such as ordering bias~\cite{pmlr-v139-zhao21c} and inherent negative bias.  
We then divide the evaluation set into three subsets based on the model’s knowledge state:  cases where the model possesses correct knowledge (\textit{parametric}), incorrect knowledge (\textit{counter-parametric}), and insufficient relevant knowledge (\textit{absent}) and convert the categorized samples into a yes-no QA and multiple-choice QA formats.

In our experiments, we find that negative bias is most pronounced in the \textit{absent} subset, where the model lacks relevant knowledge to answer the question.
That is, the negative bias problem is a type of shortcut where LLMs output negative responses when their parametric knowledge is insufficient to provide the answer.
Additionally, we analyze the impact of three prompting variants related to parametric knowledge on negative bias: the presence of context, the inclusion of an ``I don't know'' (IDK) option, and chain-of-thought (CoT) prompting ~\cite{kojima2022large}.  
Our findings suggest that while providing context and the IDK option can partially mitigate negative bias, further methodological advancements are necessary.
We also demonstrate that negative bias in binary decision tasks is strongly influenced by the prompt format.  
Specifically, instead of prompting the model to generate \textit{Yes} or \textit{No}, prompting it to select from corresponding options leads to improvements in both negative bias and weighted F1 score. 
This observation supports our claim that negative bias arises from a format-level phenomenon rather than a semantic-level one.

The contributions of our work are summarized as follows:
\begin{itemize}
    \item We show that LLMs exhibit format-level negative bias, which is amplified when the model's parametric knowledge is insufficient.
    \item We analyze the effectiveness and limitations of prompting strategies related to parametric knowledge, including providing context, the IDK option, and chain-of-thought prompting.
    \item We demonstrate that negative bias is highly sensitive to prompt type, and that converting a yes-no QA into a simple multiple-choice QA format can significantly reduce negative bias.
\end{itemize}

From the perspective of parametric knowledge, we investigate the factors influencing negative bias, with the expectation of providing insights for future research on the model's problematic behavior.

\section{Related Work}

\subsection{Hallucination Problem and Parametric Knowledge}
Hallucinations in LLMs are known to occur due to multiple factors during both the training and inference stages \cite{zhang2023siren}.
The parametric knowledge embedded in LLMs during training is closely related to the hallucination problem. 
Meng et al. \cite{NEURIPS2022_6f1d43d5} and Chaeng et al. \cite{cheang2023lms} focus on hallucinations arising from incorrect knowledge within the language model, while Wu et al. \cite{wu2024faithful} and Xie et al. \cite{xie2024adaptive} focus on hallucinations resulting from knowledge conflicts between the input context and the parametric knowledge.

In this paper, we investigate the manifestation of negative bias from the perspective of parametric knowledge. Specifically, we focus on how the presence or absence of parametric knowledge related to a given query influences the expression of negative bias, and how different prompting strategies and response formats, designed to elicit parametric knowledge, are associated with the negative bias.

\subsection{Confidence in LLM-Generated Content}
Model calibration, which refers to aligning the accuracy of the generated content with the model's confidence, plays a crucial role in assessing the reliability of the model \cite{kadavath2022language}. 
The better a model is calibrated, the higher the reliability of its generated content. 
Poorly calibrated models tend to generate incorrect content with overconfident predictions.
Recent studies have highlighted the calibration issues of LLMs \cite{zhu2023calibration} and have attempted to address these issues using approaches such as in-context learning and fine-tuning \cite{kadavath2022language,zhang2024calibrating,pmlr-v139-zhao21c}.

Negative bias is associated with the model's overconfidence in generating negative responses \cite{yu2024correcting}. Our study is closely related to research on model calibration, as we analyze the discrepancy between LLMs’ actual knowledge and their generated behavior from multiple perspectives, particularly in the context of negative responses.

\subsection{Intrinsic Bias of LLM}
LLMs have been reported to exhibit several intrinsic biases. 
For example, the lost in the middle problem, where the model's performance degrades when the relevant context is placed in the middle of the input, is observed across various LLMs \cite{liu2024lost}.

Additionally, recent studies have identified that LLMs exhibit intrinsic biases in decision-making tasks \cite{turpin2023language}. 
For example, Zheng et al. \cite{zheng2024large} demonstrate that LLMs show bias towards specific option IDs in multiple-choice question answering.

In this work, we focus on LLMs' tendency to favor negative responses over positive ones in binary decision tasks \cite{song2024large,Song2024ICiD,yu2024correcting}. 
Additionally, we aim to identify the factors contributing to this phenomenon from various perspectives, such as the parametric knowledge, prompting, and response format.

\section{Negative Bias}\label{sec:negative_bias}
\subsection{Our Approach}
In prior work, \textit{negative bias} is defined as a phenomenon observed in binary decision tasks requiring complex reasoning, where models exhibit overconfidence in negative responses, thereby producing false negatives more frequently than false positives~\cite{yu2024correcting}.
However, the current definition of negative bias fails to clarify whether the model’s bias stems primarily from the semantic content of negative responses or merely their surface format. For instance, consider the yes-no binary decision question: ``Say yes or no. Is 1+1 equal to 2?'' If the model responds with \textit{No}, this negative bias can be interpreted in two distinct ways. First, the model may believe that 1+1 equals something other than 2, indicating a semantic-level negative bias based on the contextual meaning of the answer. Alternatively, the model may simply exhibit a preference for producing the response \textit{No} regardless of semantic correctness, reflecting a format-level negative bias.

\begin{figure*}[!t]
\centering
\includegraphics[width=7.1in]{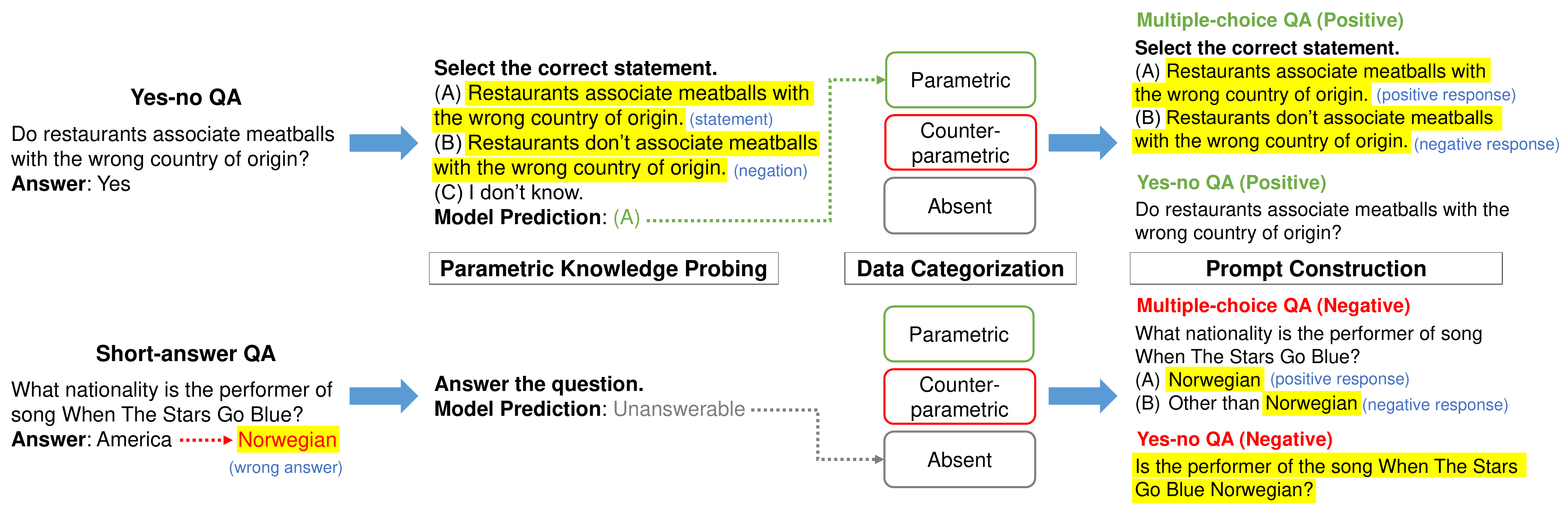}
\caption{Overview of the evaluation set construction pipeline. Examples are sampled from StrategyQA \cite{geva2021did} and 2WikiMultiHopQA \cite{ho2020constructing}. 
We highlight contents that are generated by GPT-4o \cite{achiam2023gpt}.}
\label{fig_1}
\end{figure*}

Building upon this rationale, we formulate negative bias in terms of the model’s preference to the format of the negative responses, format-level negative bias. Specifically, to observe the model’s negative bias, we compare the differences in its responses to identical binary decision questions presented in two distinct prompt formats:
\begin{itemize}
    \item \textbf{Multiple-choice QA (MCQA)}: A prompt that presents two answer options corresponding to \textit{Yes} and \textit{No} in the yes-no QA format, and asks the model to select one. 
    Notably, selecting the option corresponding to \textit{No} indicates that the model is not explicitly generating a negative response.
    \item \textbf{Yes-no QA (YNQA)}: A prompt that instructs the model to answer a binary decision question with either \textit{Yes} or \textit{No}.  
    In this prompt, we refer to responses of \textit{No} as negative responses.
\end{itemize}
Examples of each prompt type can be found in the ``Prompt Construction'' section (right) of Figure \ref{fig_1}. Unlike the YNQA type, the MCQA type requires the model to select the correct option from a list, where each option contains more specific content than a simple yes or no. Therefore, a negative response in the MCQA type takes a more indirect format compared to that in the YNQA type. In other words, if the model exhibits a negative bias in the format of its responses during binary decision tasks, this preference for negative responses is expected to be more pronounced in the YNQA type than in the MCQA type.

\subsection{Empirical Study}
We hypothesize that if a model exhibits negative bias, the proportion of negative responses in the YNQA type is higher than the proportion of selections corresponding to \textit{No} in the MCQA type. 
To validate this hypothesis, we conduct an empirical study. Specifically, we transform each sample from MuSiQue \cite{trivedi2022musique}, the short-answer multi-hop QA benchmark, into two different types: MCQA and YNQA, and then measure the degree of preference for negative responses exhibited by each type. The details of the transformation process are described in Section \ref{sec:evaluation_set_construction}. Our study is conducted on four LLMs: Llama-3.1-8B-Instruct (Llama) \cite{dubey2024llama}, Qwen2.5-7B-Instruct (Qwen) \cite{yang2024qwen2}, Mistral-7B-Instruct-v0.3 (Mistral) \cite{jiang2023mistral}, and GPT-4o-2024-08-06 (GPT-4o) \cite{achiam2023gpt}.

For each QA type, we formulate the model’s preference for negative responses as follows. Given a binary decision dataset $D$, we define the subset with the label \textit{Yes} and \textit{No} as the positive subset $D_p$ and the negative subset $D_n$, respectively.
To approximate the model $\phi$'s preference for negative responses over positive responses, we define the difference in accuracy between the negative and positive subsets as the $\Delta$:
\begin{equation}
    \Delta(D, \phi) := \text{Acc}(D_n, \phi) - \text{Acc}(D_p, \phi).
\end{equation}
A larger $\Delta$ indicates a stronger preference of the model for responses corresponding to \textit{No} in the dataset $D$.

\begin{figure}[!t]
\centering
\includegraphics[width=3.2in]{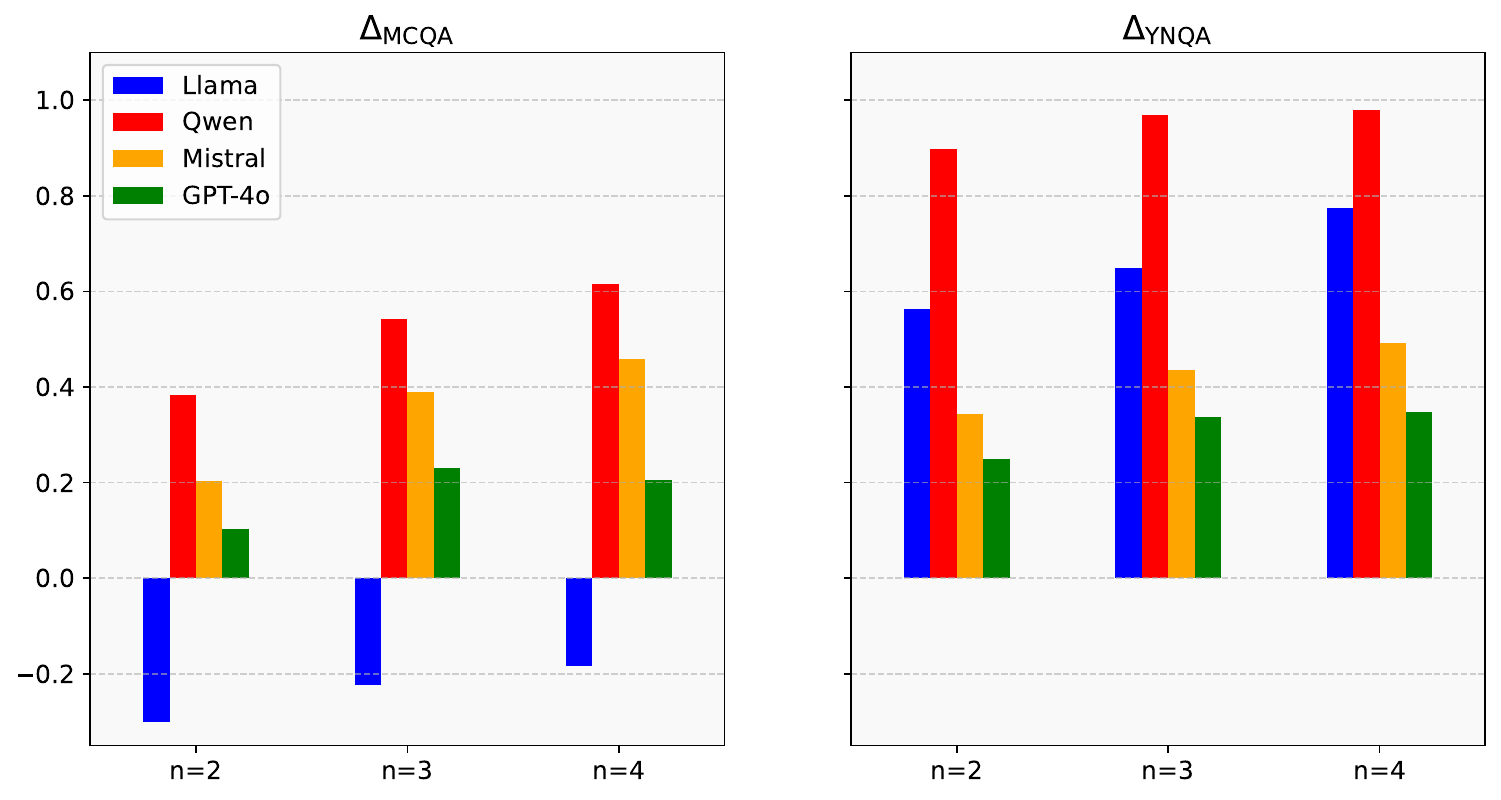}
\caption{$\Delta_{\text{MCQA}}$ and $\Delta_{\text{YNQA}}$ as a function of the number of required reasoning steps in the MCQA and YNQA formats, respectively.  
Higher values indicate that accuracy on negative samples exceeds that on positive samples. $n$ denotes the type of QA sample according to the number of hops.}
\label{fig:query_analysis}
\end{figure}

As shown in Figure \ref{fig:query_analysis}, the results of our empirical study reveal that the $\Delta$ observed in the MCQA type is consistently smaller than that of the YNQA type. Given that the two response types convey equivalent semantic content and differ only in format, this result demonstrates that LLMs’ preference for negative responses is more closely related to the format of the response than to its content.

\subsection{Definition of Negative Bias}
Based on the preceding empirical study, we observe that the negative bias exhibited by LLMs is more closely related to the format of negative responses than to their contextual content. Accordingly, we define negative bias as follows:
\textit{"A phenomenon in which, despite asking the same underlying knowledge, questions framed as yes-no question answering elicit a stronger tendency toward negative responses."}  
In other words, negative bias refers to the model's tendency to focus on the negation form of a response rather than its underlying semantic meaning.

As a proxy for negative bias, we denote negative bias score (NBS) as the difference of $\Delta$ between YNQA and MCQA types:
\begin{equation}
    \text{NBS}(D, \phi) := 0.5*\{\Delta_{\text{YNQA}}(D, \phi) - \Delta_{\text{MCQA}}(D, \phi)\}.
\end{equation}
As the NBS approaches $1$, it indicates that the model prefers responses corresponding to \textit{No} in the YNQA type more than in the MCQA type, implying a strong negative bias—that is, a tendency toward the format of negative responses.

\section{Evaluation Set Construction Pipeline}\label{sec:evaluation_set_construction}

In this section, we introduce a data splitting pipeline based on the parametric knowledge state of LLMs. 
For our experiments, we utilize open-source datasets from two categories: \textit{yes-no QA} and \textit{short-answer QA}. 
We design the pipeline considering the distinct characteristics of these dataset categories.
Our proposed pipeline consists of three sequential components: parametric knowledge probing, data categorization, and prompt construction.

\subsection{Parametric Knowledge Probing}
In this stage, we design prompts to probe the precise parametric knowledge of the LLM for each question, considering several key factors:
1) we employ CoT prompting to lead the model to predict answers robustly across the required reasoning steps,
2) to ensure that the model's predictions rely solely on its internal knowledge, we do not provide any relevant context, 
3) to account for cases where the model lacks relevant knowledge, we instruct the model to provide an IDK option to distinct response when it does not know the answer, and
4) we tailor the prompt design to align with the characteristics of each dataset type.

\subsubsection{Yes-no QA Datasets}
For yes-no QA datasets, we design a pipeline to prevent incorrect measurement of parametric knowledge caused by negative bias. 
We instruct the model to choose one of three options: a declarative statement formed from the question, the negation of that statement, or ``I don't know'' inspired by Song \cite{Song2024ICiD}. 
If the model knows the answer, it should select one of the first two options, whereas if it does not know the answer, it should select the last option.

Zhao et al. \cite{pmlr-v139-zhao21c} report that the ordering of options could affect the model's response. 
To mitigate this issue, we shuffle the order of the options three times for each sample and include only those samples in the evaluation set where the model's responses are consistent across all three shuffles.

\subsubsection{Short-answer QA Datasets}
For short-answer QA datasets, we directly ask the model to answer the question. 
The instruction includes guidance to respond with \textit{Unanswerable} if the model doesn't know the answer. 
We observe instances where the predictions do not exactly match the ground truth but appear to reflect correct parametric knowledge. 
Inspired by previous work, we use GPT-4o \cite{achiam2023gpt} to verify whether the prediction is consistent with the ground truth \cite{yu2024correcting, yu2024unleashing}.

\subsection{Data Categorization}
Based on the predictions obtained from the previous stage, we divide each dataset into three subsets based on the parametric knowledge status: correct knowledge (\textit{parametric}), incorrect knowledge (\textit{counter-parametric}), and no relevant knowledge (\textit{absent}) for each model. 

For yes-no QA datasets, we categorize each sample according to the selected option.
Specifically, if the model's response is correct, incorrect, or ``I don't know'', the sample is categorized as \textit{parametric}, \textit{counter-parametric}, or \textit{absent}, respectively.
In short-answer QA datasets, similarly, predicted answers are assigned to either \textit{parametric} or \textit{counter-parametric} subsets based on whether they align or misalign with the ground truth, respectively. 
If the model outputs \textit{Unanswerable}, the sample is categorized into the \textit{absent} subset.
To ensure clarity in observation, we use the single word \textit{Unanswerable} instead of the phrase ``I don't know'' in the instruction for short-answer QA datasets.

\begin{figure*}[!t]
\centering
\includegraphics[width=7.2in]{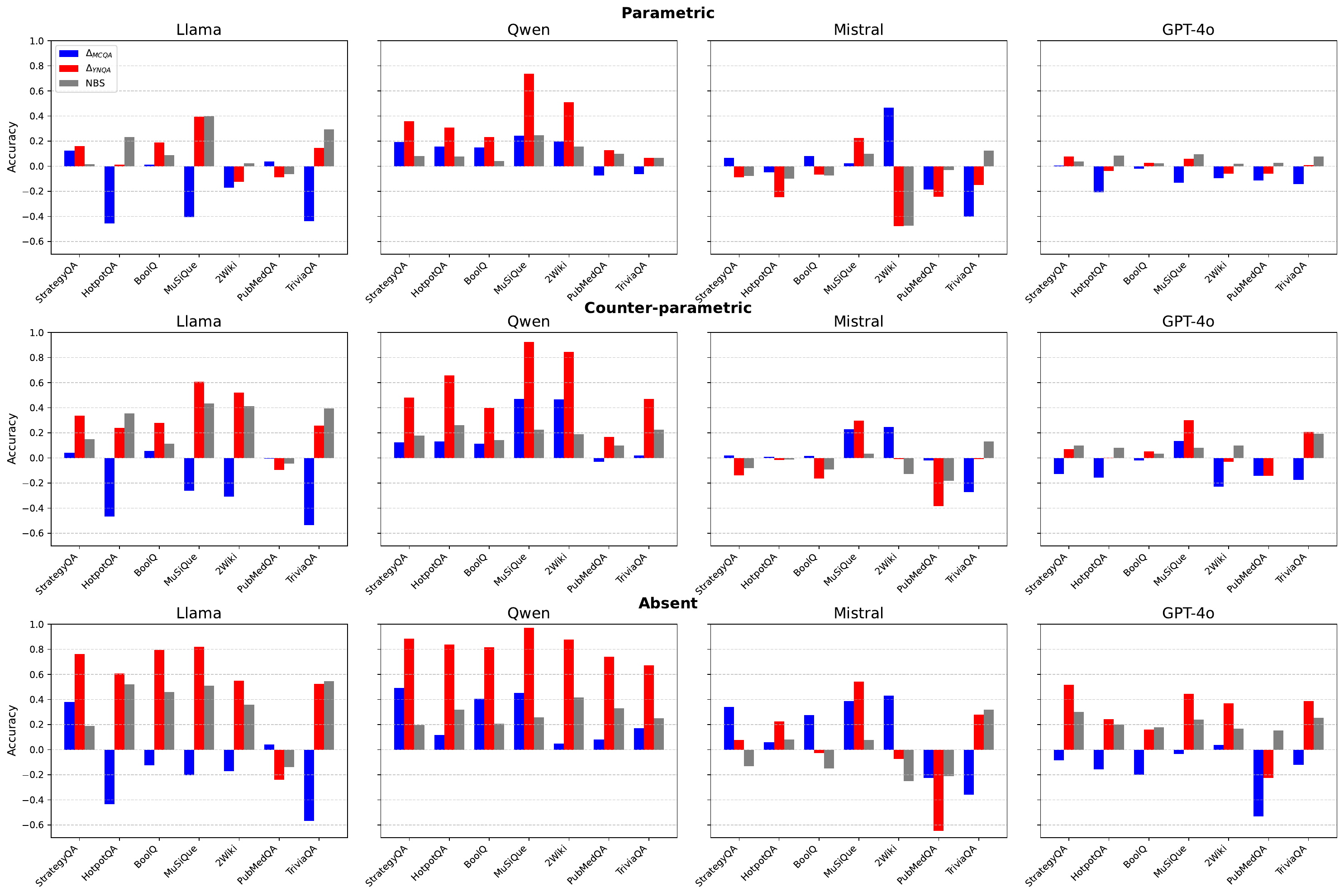}
\caption{The results for $\Delta_{\text{MCQA}}$, $\Delta_{\text{YNQA}}$ and NBS. For brevity, 2WikiMultiHopQA is abbreviated as 2Wiki in the figure.}
\label{fig:default_acc}
\end{figure*}

\subsection{Prompt Construction}
As a final step, each sample is converted into either the MCQA or YNQA type.  
We design the two provided options in the MCQA type to be semantically aligned to \textit{Yes} and \textit{No} in the YNQA type.

\noindent\textbf{MCQA Type}  
For yes-no QA datasets, we leverage the statement–negation pairs generated during the parametric knowledge probing process.  
Each sample is mapped to either a positive or negative example according to its label.  
For short-answer QA datasets, we construct positive or negative samples using either the correct label or a generated incorrect label, respectively.

\noindent\textbf{YNQA Type} 
For yes-no QA datasets, we retain the original questions without any modification, and the ground truth is used as is.
For short-answer QA datasets, we balance the number of samples where the label corresponds to \textit{Yes} (i.e., positive samples) and \textit{No} (i.e., negative samples). 
For positive samples, we use GPT-4o to generate yes-no questions based on the ground truth and the original question. 
For negative samples, we first use GPT-4o to generate the incorrect answer by considering the context, the question, and the ground truth. 
Note that we iterate the incorrect answer generation process until the generated label is different from the model's prediction.
We then generate negative samples using GPT-4o, utilizing the generated incorrect answer and the original question, similar to the case of positive samples.

As a result, each model has seven evaluation sets, with each evaluation set divided into three subsets based on the model's parametric knowledge state. 
Furthermore, each subset contains both positive and negative samples, which we call positive and negative subsets, respectively.
\section{Experimental Setup}\label{sec:experimental_setup}

\subsection{Datasets}
We utilize three yes-no QA datasets: StrategyQA \cite{geva2021did}, BoolQ \footnote[1]{\url{https://huggingface.co/datasets/google/boolq}} \cite{clark2019boolq}, and PubMedQA \footnote[2]{\url{https://huggingface.co/datasets/qiaojin/PubMedQA}} \cite{jin2019pubmedqa}, along with four short-answer QA datasets: HotpotQA \footnote[3]{\url{https://huggingface.co/datasets/hotpotqa/hotpot_qa}} \cite{yang2018hotpotqa}, MuSiQue \footnote[4]{\url{https://huggingface.co/datasets/bdsaglam/musique}}, 2WikiMultiHopQA \footnote[5]{\url{https://huggingface.co/datasets/voidful/2WikiMultihopQA}}, and TriviaQA \footnote[6]{\url{https://huggingface.co/datasets/mandarjoshi/trivia_qa}} \cite{joshi2017triviaqa}. 
All datasets include annotated contexts required for inference. 
For StrategyQA, we concatenate the supporting facts to form the context. 
In the case of MuSiQue, we use only the paragraphs containing supporting facts as the context. 
For the remaining datasets, we use the entire annotated context.

We observe that, in some datasets, there are very few positive or negative samples within certain subsets. 
If a subset contains fewer than 50 positive or negative samples, and additional sources are available within the dataset (e.g., training set), we augment the subset with up to 50 samples. 
Statistics for the evaluation sets of each model and details of sources for datasets are shown in Table \ref{tab:statistics} and Appendix \ref{app:details_of_dataset}, respectively.

\subsection{Models} 
We employ four LLMs: Llama-3.1-8B-Instruct (Llama) \cite{dubey2024llama}, Qwen2.5-7B-Instruct (Qwen) \cite{yang2024qwen2}, Mistral-7B-Instruct-v0.3 (Mistral) \cite{jiang2023mistral}, and GPT-4o-2024-08-06 (GPT-4o) \cite{achiam2023gpt} as models for analysis. 
For the first three LLMs, we further conduct an analysis of negative attention score in Section \ref{sec:negative_attention_score_analysis}. 
Experiments utilizing Llama, Qwen, and Mistral are conducted using the \textit{HuggingFace Transformers}\footnote[7]{\url{https://github.com/huggingface/transformers}} \cite{wolf2020transformers}.

\subsection{Metrics}\label{subsec:metrics}
To analyze the model's negative bias in terms of accuracy and calibration, we use the following metrics:
\begin{itemize}
\item \textbf{$\Delta_{\text{MCQA}}$ / $\Delta_{\text{YNQA}}$:} The tendency of the model to generate negative responses in the MCQA or YNQA format.
\item \textbf{Negative Bias Score (NBS):} The degree to which the model is more inclined to return negative responses in the YNQA format compared to the MCQA format.
\item \textbf{Weighted F1 Score:} To evaluate overall performance on the binary decision task, we report the weighted F1 score in Appendices~\ref{app:default_raw_results}, due to the imbalance between positive and negative samples in the yes-no QA datasets.  
Note that only samples yielding either a positive or negative response are included in the evaluation.
\end{itemize}
\section{Initial Observation}\label{sec:initial_observation}

We measured $\Delta_{\text{MCQA}}$, $\Delta_{\text{YNQA}}$, and NBS for four LLMs across seven datasets, as shown in Figure~\ref{fig:default_acc}. 
The $\Delta$ values for both types, along with the weighted F1 scores, can be found in Table~\ref{tab:default_result} in Appendix~\ref{app:default_raw_results}.

The most important finding is that negative bias is most pronounced in the \textit{absent} subsets.  
For the \textit{absent} subsets, NBS takes a positive value in 85.7\% of the 28 cases.  
In contrast, negative bias appears relatively weaker when the LLMs possess knowledge relevant to the question (i.e., \textit{parametric} or \textit{counter-parametric} subsets).  
While $\Delta_{\text{YNQA}}$ is mostly positive in the \textit{absent} subsets, $\Delta_{\text{MCQA}}$ does not exhibit a consistent sign tendency.  
Additionally, as shown by the weighted F1 scores in Table~\ref{tab:default_result}, the MCQA type generally achieves higher scores.  
Taken together, these results suggest that even when the binary decision questions are semantically equivalent, prompting with the YNQA type leads to a higher frequency of negative responses—enough to undermine reliability compared to the MCQA type.

Another key finding is that negative bias does not show a strong correlation with model size.  
Among the three 7B models, Qwen exhibits a clear negative bias, whereas Mistral shows a relatively weak one.  
Furthermore, GPT-4o—a much larger model—exhibits stronger negative bias than Mistral.  
These results suggest that negative bias is nearly orthogonal to model size and should be treated as an independent issue.

Based on these observations, we hypothesize the following:  
\textit{When a model lacks sufficient knowledge required to answer a yes-no question, there exists a shortcut tendency to respond negatively.}  
To further analyze the relationship between parametric knowledge and negative bias, we formulate three research questions:

\noindent\textbf{RQ1: Does introducing external knowledge reduce negative bias?}  
This setting provides input containing the information required to answer the question correctly, resembling a retrieval-augmented generation scenario.  
We investigate whether the aforementioned shortcut tendency is mitigated when the missing knowledge is explicitly provided as input.  
In addition, we examine the case where the external knowledge conflicts with the model’s parametric knowledge using the \textit{counter-parametric} subsets.

\noindent\textbf{RQ2: Does including an ``I don't know'' option in the prompt reduce the frequency of negative responses?}  
We observed that the YNQA type induces an excessive number of negative responses compared to the MCQA type.  
This aligns with the findings of Yu et al.~\cite{yu2024correcting} that LLMs tend to produce overconfident negative responses in YNQA settings.  
We further observe whether adding an IDK option beyond binary choices can improve model calibration and thereby alleviate negative bias.

\noindent\textbf{RQ3: How does chain-of-thought prompting affect negative bias?}  
CoT prompting is a simple yet effective method for enhancing a model’s reasoning capabilities.  
We explore whether strengthening reasoning through CoT prompting helps mitigate the shortcut behavior that leads to negative responses.

By examining how negative bias changes under these three general prompting scenarios, we aim to identify effective components for mitigating negative bias and to analyze their limitations.

\begin{figure*}[!t]
\centering
\includegraphics[width=6.5in]{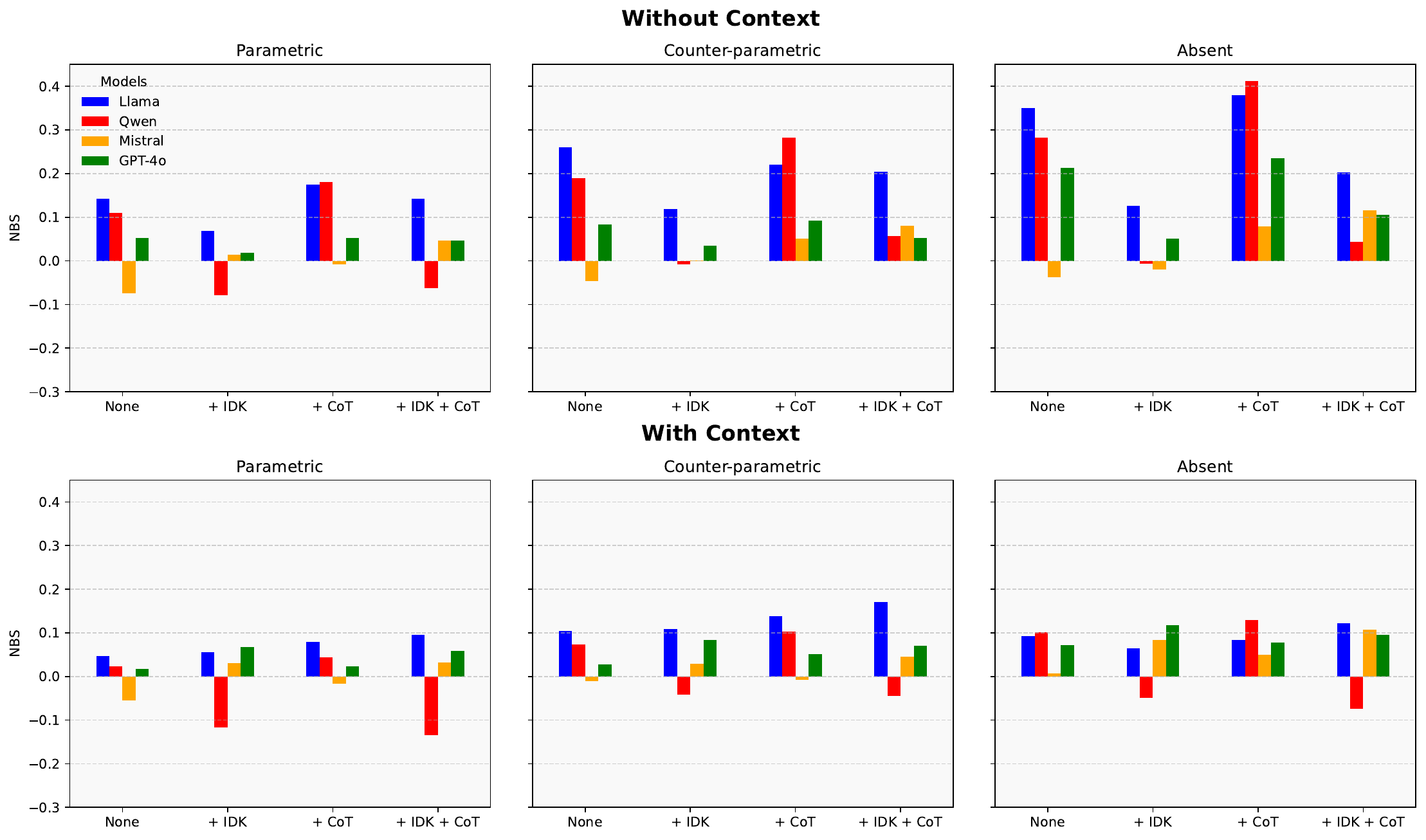}
\caption{The NBS results averaged across datasets with and without the application of three prompting criteria.}
\label{fig:prompting_scenario}
\end{figure*}

\section{Negative Bias on Various Prompting Scenarios}\label{sec:prompting_scenario}
To obtain answers to the research questions, we set up eight prompting scenarios based on three orthogonal criteria:

\begin{itemize}
\item \textbf{Context:} The ``with context'' scenario resembles a reading comprehension task, where the relevant context necessary for answering the question is provided. 
Note that a knowledge conflict arises in the \textit{counter-parametric} subset. 
The ``without context'' scenario represents a closed-book QA setting where only an instruction and a question are provided as input. 
In this setting, the model responds based solely on the parametric knowledge.
\item \textbf{IDK Option:} In the ``with IDK option'' scenario, the prompt includes an option for the model to respond with \textit{Unanswerable} if it lacks sufficient knowledge to provide an answer. 
In the ``without IDK option'' scenario, the model is restricted to choosing only between \textit{Yes} and \textit{No}. 
\item \textbf{CoT Prompting:} For the ``with chain-of-thought prompting'' scenario, we adopt the zero-shot CoT prompting \cite{kojima2022large}.
Our objective is to determine whether CoT prompting amplifies or mitigates the model's negative bias.
\end{itemize}
We note that the \textit{default} prompt scenario refers to the prompting scenario where none of the three criteria are applied.
Figure~\ref{fig:prompting_scenario} presents the average NBS of the four models under eight prompting scenarios.

\begin{table*}
\centering
\caption{The ratio of predictions that shifted to IDK after applying the IDK option. The gray-colored column indicates cases where correct predictions are shifted to IDK.}\label{tab:prediction_shift}
\resizebox{0.88\linewidth}{!}{
\begin{tabular}{l|>{\columncolor[gray]{0.8}}ccc>{\columncolor[gray]{0.8}}c|>{\columncolor[gray]{0.8}}ccc>{\columncolor[gray]{0.8}}c|>{\columncolor[gray]{0.8}}ccc>{\columncolor[gray]{0.8}}c}
\hline
\toprule
\multicolumn{1}{c|}{ } &
\multicolumn{4}{c|}{Parametric} &
\multicolumn{4}{c|}{Counter-parametric} & 
\multicolumn{4}{c}{Absent} \\
\multicolumn{1}{c|}{ } &
\multicolumn{2}{c}{Positive} &
\multicolumn{2}{c|}{Negative} & 
\multicolumn{2}{c}{Positive} &
\multicolumn{2}{c|}{Negative} & 
\multicolumn{2}{c}{Positive} &
\multicolumn{2}{c}{Negative} \\
Model & Yes$\rightarrow$ & No$\rightarrow$ & Yes$\rightarrow$ & No$\rightarrow$ & Yes$\rightarrow$ & No$\rightarrow$ & Yes$\rightarrow$ & No$\rightarrow$ & Yes$\rightarrow$ & No$\rightarrow$ & Yes$\rightarrow$ & No$\rightarrow$ \\
\hline
\multicolumn{13}{c}{\textit{MCQA}} \\
\hline
Llama & 0.001 & 0.000 & 0.006 & 0.010 & 0.005 & 0.005 & 0.003 & 0.060 & 0.016 & 0.099 & 0.032 & 0.123 \\
Qwen & 0.021 & 0.168 & 0.073 & 0.088 & 0.095 & 0.218 & 0.067 & 0.396 & 0.177 & 0.600 & 0.205 & 0.519 \\
Mistral & 0.010 & 0.284 & 0.020 & 0.094 & 0.017 & 0.190 & 0.015 & 0.256 & 0.065 & 0.453 & 0.110 & 0.406 \\
GPT-4o & 0.016 & 0.182 & 0.100 & 0.047 & 0.047 & 0.086 & 0.060 & 0.173 & 0.339 & 0.633 & 0.428 & 0.521 \\
\hline
\multicolumn{13}{c}{\textit{YNQA}} \\
\hline
Llama & 0.074 & 0.305 & 0.223 & 0.254 & 0.132 & 0.458 & 0.091 & 0.474 & 0.427 & 0.785 & 0.667 & 0.783 \\
Qwen & 0.089 & 0.693 & 0.233 & 0.598 & 0.122 & 0.776 & 0.147 & 0.840 & 0.301 & 0.938 & 0.457 & 0.939 \\
Mistral & 0.005 & 0.035 & 0.021 & 0.102 & 0.026 & 0.157 & 0.035 & 0.144 & 0.136 & 0.412 & 0.162 & 0.366 \\
GPT-4o & 0.032 & 0.162 & 0.082 & 0.072 & 0.061 & 0.166 & 0.099 & 0.140 & 0.336 & 0.617 & 0.443 & 0.554 \\

\bottomrule
\end{tabular}}
\end{table*}

\subsection{External Knowledge}
Comparing the first and second rows of Figure~\ref{fig:prompting_scenario}, we observe that providing relevant context reduces NBS.  
With the inclusion of context, the gap in NBS among the three subsets categorized by parametric knowledge status becomes smaller.  
This highlights that negative bias is strongly related to the parametric knowledge status.

However, there are limitations to simply providing context.  
Compared to the \textit{parametric} subset, the relatively higher NBS observed in the \textit{counter-parametric} and \textit{absent} subsets remains.  
This suggests that knowledge conflict has not been fully resolved, or that knowledge injection is insufficient to address the issue.  
It implies that grounding external knowledge in the model requires additional mechanisms in order to effectively mitigate negative bias.

\subsection{IDK Option}
In most cases, providing the model with an IDK option (i.e., +IDK or +IDK+CoT in Figure~\ref{fig:prompting_scenario}) leads to a decreasing trend in NBS.  
As shown in Table~\ref{tab:prompting_scenario_result} in Appendix~\ref{app:default_raw_results}, the inclusion of the IDK option substantially reduces $\Delta_{\text{YNQA}}$ compared to $\Delta_{\text{MCQA}}$.  
This indicates a significant decline in negative responses relative to positive ones, suggesting that the model's overconfidence in negative answers has been alleviated through improved calibration.

\begin{figure*}[!t]
\centering
\includegraphics[width=6.6in]{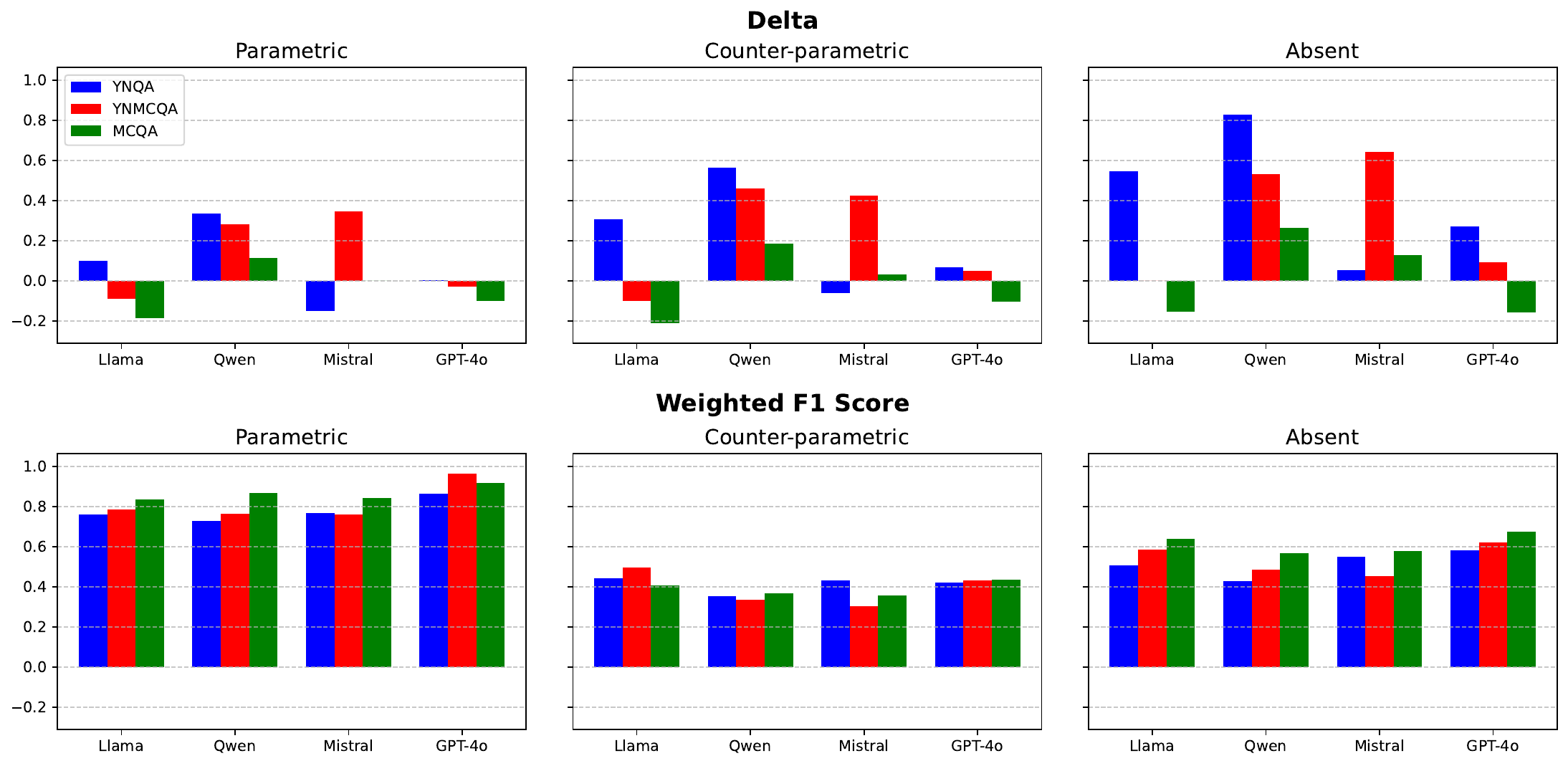}
\caption{$\Delta$ (first row) and weighted F1 score (second row) across different prompt types.}
\label{fig:format_analysis}
\end{figure*}

We further analyze how the presence of the IDK option influences model predictions. 
Table~\ref{tab:prediction_shift} presents the proportion of responses across the seven datasets that shift to the IDK option from either a positive response (i.e., Yes$\rightarrow$) or a negative response (i.e., No$\rightarrow$) when the IDK option is available.  
Note that positive and negative responses in the MCQA type correspond to the selection of semantically aligned options with \textit{Yes} or \textit{No} in the YNQA type.  
The columns highlighted in gray indicate the proportion of originally correct predictions that shift to IDK.
Based on this, we identify two major findings.

\noindent\textbf{Negative responses are more likely to shift to IDK.}  
In all cases, the proportion of negative responses shifting to IDK (i.e., No$\rightarrow$) is higher than that of positive responses (i.e., Yes$\rightarrow$).  
We also observe that the prediction shift is more frequent in the YNQA type than in the MCQA type, which leads to a reduction in NBS.  
Moreover, the \textit{absent} subsets exhibit a higher shift rate than the other subsets.  
Taken together, these findings suggest that the IDK option enhances model calibration and mitigates the tendency to produce unwarranted negative responses.

\noindent\textbf{The IDK option may overly suppress both positive and negative responses.}  
In the \textit{parametric} subset, we find cases where predictions that were originally correct shift to IDK, especially from negative responses.  
The IDK option reduces not only false negatives but also true negatives, which in turn leads to a decrease in the weighted F1 score in the \textit{parametric} subset, as shown in Table~\ref{tab:prompting_scenario_result}.  
These findings imply that while the IDK option helps mitigate negative bias, further research is needed to develop methods that improve model calibration in a more fine-grained manner.

\subsection{Chain-of-Thought}
In Figure~\ref{fig:prompting_scenario}, CoT prompting (i.e., +CoT or +IDK+CoT) leads to an increase in NBS in most cases.  
Although CoT is expected to elicit the model's parametric knowledge and thereby mitigate negative bias, it appears that internal bias is amplified during the reasoning process.  
As shown in Table~\ref{tab:prompting_scenario_result}, the MCQA type consistently exhibits an increase in weighted F1 score when CoT is applied.  
In contrast, the YNQA type shows only marginal improvements—or even decreases—in performance.  
This suggests that CoT substantially alters the response distribution, specifically in the YNQA type, which in turn enhances negative bias.

\section{Prompt Type Analysis}\label{sec:prompt_format_analysis}

We have demonstrated that the absence of parametric knowledge amplifies negative bias.  
Additionally, by comparing the MCQA and YNQA types, we have shown that prompting in a way that avoids explicitly generating negative words (i.e., \textit{No}) can mitigate the model's tendency to produce negative responses.  
In this section, we compare the $\Delta$ and weighted F1 scores across different prompt types.

In addition to YNQA and MCQA, we define a third prompt type: yes-no multiple-choice QA (YNMCQA), in which the binary decision question is followed by multiple-choice options labeled \textit{Yes} and \textit{No}.

\begin{tcolorbox}[colback=gray!5!white, colframe=black, 
  arc=4mm, boxrule=0.8pt, left=2mm, right=2mm, top=1mm, bottom=1mm]
\textbf{Question:} Do restaurants associate meatballs with the wrong country of origin?\\
\textbf{Options:} (A) Yes (B) No
\end{tcolorbox}

Note that, unlike the MCQA type, this approach does not require generating content-specific options for the binary decision question.

Figure~\ref{fig:format_analysis} presents the $\Delta$ values and weighted F1 scores across the three prompt types.  
One notable finding is that, except for Mistral, converting to the YNMCQA type results in lower $\Delta$ values and higher weighted F1 scores compared to the YNQA type in most cases.  
This highlights that directly generating the negative word from the model is one of the key factors contributing to negative bias.  
In conclusion, simply altering the answer structure of yes-no QA can alleviate the model's tendency to produce excessive negative responses.

\section{Negative Attention Score Analysis}\label{sec:negative_attention_score_analysis}

To better understand the mechanisms by which prompting techniques address negative bias, we conduct an analysis of attention patterns. 
Yu et al. \cite{yu2024correcting} demonstrated that the tendency of LLMs to produce negative predictions in binary decision tasks is closely related to the attention scores assigned to negative tokens (e.g., ``No'') within the user instruction. 
They introduced the concept of the negative attention score (NAS), showing that when both positive (e.g., ``Yes'') and negative tokens are present in the user instruction, models exhibit a tendency to assign higher attention to the negative tokens.

Suppose the input prompt $x$ consists of a task instruction $\{x_1, \ldots, x_{N_I}\}$ and a user input $\{x_{N_{I+1}}, \ldots, x_N\}$. 
The NAS at the $l$-th layer and $h$-th attention head is defined as follows:
\begin{equation}
    \text{NAS}{(x, l, h)} := 
    \sum^{N}_{i=N_I} \left(A^{l,h}_{i,t_{p}} + A^{l,h}_{i,t_{n}}\right) 
    \cdot \log\left(\frac{A^{l,h}_{i,t_{n}}}{A^{l,h}_{i,t_{p}}}\right),
    \label{eq:m_cl}
\end{equation}
where $A^{l,h}_{i,t_p}$ and $A^{l,h}_{i,t_n}$ denote the attention scores that the $i$-th token assigns to the positions of the positive and negative tokens ($t_p$ and $t_n$, respectively). 
In other words, the NAS is designed to increase when the model assigns relatively greater attention to the negative token compared to the positive token in the user instruction.

To account for the influence of reasoning generated by CoT prompting on NAS, we measure the average NAS across all token positions produced until the model generates its final response. 
Specifically, let the input–output sequence prior to producing the answer be denoted as $y = \{y_1, \ldots, y_N, \ldots, y_M\}$. 
We then define the mean NAS (mNAS) as follows:
\begin{equation}
    \text{mNAS}{(y)} := \frac{1}{M}\sum_{l}{\sum_{h}{\text{NAS}(y, l, h)}}.
\end{equation}

\begin{table}[t!]
    \centering
    \caption{Results of per-token model NAS (mNAS) across different prompting scenarios.}
    {\resizebox{0.78\columnwidth}{!}{
    \renewcommand{\arraystretch}{1.1}
    \begin{tabular}{l|ccc}
    \toprule
       \textbf{Llama}   & \textbf{Parametric} & \textbf{Counter.} & \textbf{Absent}\\ \hline
       None & \textbf{5.88} & \textbf{5.74} & \textbf{5.71} \\
       $+$ IDK & 4.90 & 4.76 &  4.60 \\
    \hline
        $+$ Context & 1.56 & 1.51 & 1.63 \\
       $+$ CoT & \textbf{4.65} & \textbf{4.45} &  \textbf{4.93} \\
    \hline
       \textbf{Qwen}   & \textbf{Parametric} & \textbf{Counter.} & \textbf{Absent}\\ \hline
       None & \textbf{0.69} & \textbf{0.82} & \textbf{0.98} \\
       $+$ IDK & $-$0.14 & $-$0.17 &  $-$0.24 \\
   \hline
        $+$ Context & 0.08 & 0.03 &  0.07 \\
       $+$ CoT & \textbf{0.61} & \textbf{0.62} &  \textbf{0.72} \\
    \hline
       \textbf{Mistral}   & \textbf{Parametric} & \textbf{Counter.} & \textbf{Absent}\\ \hline
       None & 1.95 & 2.00 &  2.07 \\
       $+$ IDK & \textbf{2.21} & \textbf{2.19} &  \textbf{2.20} \\
   \hline
        $+$ Context & 0.18 & 0.15 &  0.19 \\
        $+$ CoT & \textbf{0.97} & \textbf{0.96} &  \textbf{1.04} \\
    \bottomrule
    \end{tabular}
    }}    
    \label{tab:mNAS_analysis}
\end{table}
\noindent\textbf{None vs IDK.} We first compare the cases without any additional context or reasoning process.
For Llama and Qwen, providing the IDK option in the instruction leads to a reduction in mNAS, while Mistral shows an increase.
These findings are consistent with Figure~\ref{fig:prompting_scenario}, where only Mistral shows an increase in NBS when the IDK option is introduced.

\noindent\textbf{Context vs CoT.} Compared to the provision of additional context, the CoT prompting exhibits a relatively higher mNAS. 
This aligns with the observation in Figure~\ref{fig:prompting_scenario}, where CoT prompting increases NBS. 
As an additional model input, supporting facts substantially reduce mNAS, whereas the self-reasoning process leads to only a modest reduction. 
This suggests that the accumulation of additional tokens can either decrease or increase NBS depending on how they affect attention patterns.

Our results also suggest that the negative bias analysis from the perspective of the attention mechanism is consistent with the experimental results on the relationship between prompting scenarios and NBS.
\section{Discussion}

While our study has yielded several insightful conclusions, we foresee two directions for further research that may deepen the understanding of negative bias in LLMs and contribute to its mitigation:

\noindent\textbf{Further analysis of parametric knowledge.}
As described in Section \ref{sec:evaluation_set_construction}, we introduced prompting strategies and a CoT process to minimize negative bias, thereby enabling accurate probing of parametric knowledge. 
As a result, we observed a clear tendency for negative bias to intensify in the order of \textit{absent}, \textit{counter-parametric}, and \textit{parametric} subsets, as shown in Figures \ref{fig:prompting_scenario} and \ref{fig:format_analysis}. 
However, the precise extraction of parametric knowledge from LLMs remains an ongoing research \cite{chen2024inside, simhi2025distinguishingignoranceerrorllm}. 
We expect that developing additional techniques to reduce bias will yield clearer observations.

Moreover, future research could refine a three-way data-splitting approach to enable more fine-grained analyses. 
For example, one may distinguish between samples in \textit{counter-parametric} subset that yield correct answers with context (i.e., \textit{weak counter-parametric}) versus those that do not (i.e., \textit{strong counter-parametric}), or separate samples where the IDK option leads to an \textit{Unanswerable} response versus cases where the model produces a binary decision.

\noindent\textbf{Connecting model training with negative bias.}
Our analysis primarily focused on identifying inference-time factors that influence negative bias. 
However, there remains room for exploring negative bias from the perspective of training dynamics. 
While this lies beyond the scope of the present work, future studies that examine negative bias in relation to training appear promising for informing model optimization.
\section{Conclusion}

In this paper, we defined negative bias in yes-no QA as a phenomenon in which models focus on and prefer the negation form of a response, rather than its semantic meaning.  
Through extensive experiments, we analyzed the nature of negative bias in terms of both parametric knowledge and prompt type.

We summarize the main conclusions and actionable takeaways from our study as follows.

\noindent\textbf{In yes-no QA, when the model lacks relevant knowledge, a shortcut tendency emerges toward negative bias.}  
In Section~\ref{sec:initial_observation}, we divided the datasets into three subsets based on the model's parametric knowledge status and observed that negative bias was most pronounced in the \textit{absent} subset.  
This is further supported by the reduction in NBS when relevant context is included in the prompt.

\noindent\textbf{Providing context and an IDK option mitigates negative bias, whereas CoT prompting amplifies it.}  
In Section~\ref{sec:prompting_scenario}, we investigated three approaches that aim to enhance accuracy from a parametric knowledge perspective and examined their relationship with negative bias.  
Providing context serves as a direct method of injecting necessary information for correct inference, and we observed a reduction in NBS.  
The IDK option improves model calibration and reduces inaccurate responses, thereby yielding a positive effect on negative bias.  
In contrast, CoT prompting, which is designed to elicit the model’s reasoning process, often amplifies intrinsic biases, leading to an increase in NBS.  
While we found that context and the IDK option partially address negative bias, our results indicate the need for further research into more effective methods of knowledge injection and fine-grained model calibration.

\noindent\textbf{Negative bias can be alleviated by simply preventing the model from generating negative words.}  
In Section~\ref{sec:prompt_format_analysis}, we observed that the value of $\Delta$ varies depending on the prompt design of the binary decision question.  
We found that prompting the model to generate explicit negative words during response generation contributes to negative bias.  
At the same time, simple prompt rephrasing—as in the YNMCQA format—resulted in improvements in both bias and weighted F1 score, demonstrating that careful prompt design can be an effective mitigation strategy.

We hope our study lays the groundwork for future efforts to understand and reduce model bias in binary decision tasks.

\section{Acknowledgements}
This work was supported by the BK21 FOUR program of the Education and Research Program for Future ICT Pioneers, Seoul National University, Institute of Information \& communications Technology Planning \& Evaluation (IITP) grant funded by the Korea government(MSIT) [NO. RS-2021-II211343, Artificial Intelligence Graduate School Program (Seoul National University), RS-2022-II220959, No. RS-2025-02263754, Human-Centric Embodied AI Agents with Autonomous Decision-Making, IITP-2025-RS-2024-00397085], the National Research Foundation of Korea (NRF) grant funded by the Korea government (MSIT) (No. 2022R1A3B1077720, No. 2022R1A5A708390811).

\bibliographystyle{IEEEtran}
\bibliography{custom}

@article{ouyang2022training,
  title={Training language models to follow instructions with human feedback},
  author={Ouyang, Long and Wu, Jeffrey and Jiang, Xu and Almeida, Diogo and Wainwright, Carroll and Mishkin, Pamela and Zhang, Chong and Agarwal, Sandhini and Slama, Katarina and Ray, Alex and others},
  journal={Advances in neural information processing systems},
  volume={35},
  pages={27730--27744},
  year={2022}
}

@article{dubey2024llama,
  title={The llama 3 herd of models},
  author={Dubey, Abhimanyu and Jauhri, Abhinav and Pandey, Abhinav and Kadian, Abhishek and Al-Dahle, Ahmad and Letman, Aiesha and Mathur, Akhil and Schelten, Alan and Yang, Amy and Fan, Angela and others},
  journal={arXiv preprint arXiv:2407.21783},
  year={2024}
}

@article{jiang2023mistral,
  title={Mistral 7B},
  author={Jiang, Albert Q and Sablayrolles, Alexandre and Mensch, Arthur and Bamford, Chris and Chaplot, Devendra Singh and Casas, Diego de las and Bressand, Florian and Lengyel, Gianna and Lample, Guillaume and Saulnier, Lucile and others},
  journal={arXiv preprint arXiv:2310.06825},
  year={2023}
}

@article{yang2024qwen2,
  title={Qwen2 Technical Report},
  author={Yang, An and Yang, Baosong and Hui, Binyuan and Zheng, Bo and Yu, Bowen and Zhou, Chang and Li, Chengpeng and Li, Chengyuan and Liu, Dayiheng and Huang, Fei and others},
  journal={CoRR},
  year={2024}
}

@article{achiam2023gpt,
  title={Gpt-4 technical report},
  author={Achiam, Josh and Adler, Steven and Agarwal, Sandhini and Ahmad, Lama and Akkaya, Ilge and Aleman, Florencia Leoni and Almeida, Diogo and Altenschmidt, Janko and Altman, Sam and Anadkat, Shyamal and others},
  journal={arXiv preprint arXiv:2303.08774},
  year={2023}
}

@article{kojima2022large,
  title={Large language models are zero-shot reasoners},
  author={Kojima, Takeshi and Gu, Shixiang Shane and Reid, Machel and Matsuo, Yutaka and Iwasawa, Yusuke},
  journal={Advances in neural information processing systems},
  volume={35},
  pages={22199--22213},
  year={2022}
}

@article{madaan2024self,
  title={Self-refine: Iterative refinement with self-feedback},
  author={Madaan, Aman and Tandon, Niket and Gupta, Prakhar and Hallinan, Skyler and Gao, Luyu and Wiegreffe, Sarah and Alon, Uri and Dziri, Nouha and Prabhumoye, Shrimai and Yang, Yiming and others},
  journal={Advances in Neural Information Processing Systems},
  volume={36},
  year={2024}
}

@inproceedings{
wei2022finetuned,
title={Finetuned Language Models are Zero-Shot Learners},
author={Jason Wei and Maarten Bosma and Vincent Zhao and Kelvin Guu and Adams Wei Yu and Brian Lester and Nan Du and Andrew M. Dai and Quoc V Le},
booktitle={International Conference on Learning Representations},
year={2022},
url={https://openreview.net/forum?id=gEZrGCozdqR}
}

@article{brown2020language,
  title={Language models are few-shot learners},
  author={Brown, Tom B},
  journal={arXiv preprint arXiv:2005.14165},
  year={2020}
}

@article{zhang2023siren,
  title={Siren's song in the AI ocean: a survey on hallucination in large language models},
  author={Zhang, Yue and Li, Yafu and Cui, Leyang and Cai, Deng and Liu, Lemao and Fu, Tingchen and Huang, Xinting and Zhao, Enbo and Zhang, Yu and Chen, Yulong and others},
  journal={arXiv preprint arXiv:2309.01219},
  year={2023}
}

@article{kaufmann2023survey,
  title={A survey of reinforcement learning from human feedback},
  author={Kaufmann, Timo and Weng, Paul and Bengs, Viktor and H{\"u}llermeier, Eyke},
  journal={arXiv preprint arXiv:2312.14925},
  year={2023}
}

@inproceedings{
xie2024adaptive,
title={Adaptive Chameleon  or Stubborn Sloth: Revealing the Behavior of Large Language Models in Knowledge Conflicts},
author={Jian Xie and Kai Zhang and Jiangjie Chen and Renze Lou and Yu Su},
booktitle={The Twelfth International Conference on Learning Representations},
year={2024},
url={https://openreview.net/forum?id=auKAUJZMO6}
}

@inproceedings{addlesee2024grounding,
    title = "Grounding {LLM}s to In-prompt Instructions: Reducing Hallucinations Caused by Static Pre-training Knowledge",
    author = "Addlesee, Angus",
    editor = "Dinkar, Tanvi  and
      Attanasio, Giuseppe  and
      Cercas Curry, Amanda  and
      Konstas, Ioannis  and
      Hovy, Dirk  and
      Rieser, Verena",
    booktitle = "Proceedings of Safety4ConvAI: The Third Workshop on Safety for Conversational AI @ LREC-COLING 2024",
    month = may,
    year = "2024",
    address = "Torino, Italia",
    publisher = "ELRA and ICCL",
    url = "https://aclanthology.org/2024.safety4convai-1.1",
    pages = "1--7",
}

@inproceedings{
su2024textttconflictbank,
title={\${\textbackslash}texttt\{ConflictBank\}\$: A Benchmark for Evaluating the Influence of Knowledge Conflicts in {LLM}s},
author={Zhaochen Su and Jun Zhang and Xiaoye Qu and Tong Zhu and Yanshu Li and Jiashuo Sun and Juntao Li and Min Zhang and Yu Cheng},
booktitle={The Thirty-eight Conference on Neural Information Processing Systems Datasets and Benchmarks Track},
year={2024},
url={https://openreview.net/forum?id=wjHVmgBDzc}
}

@article{yu2024correcting,
  title={Correcting negative bias in large language models through negative attention score alignment},
  author={Yu, Sangwon and Song, Jongyoon and Hwang, Bongkyu and Kang, Hoyoung and Cho, Sooah and Choi, Junhwa and Joe, Seongho and Lee, Taehee and Gwon, Youngjune L and Yoon, Sungroh},
  journal={arXiv preprint arXiv:2408.00137},
  year={2024}
}

@inproceedings{cheang2023lms,
    title = "Can {LM}s Generalize to Future Data? An Empirical Analysis on Text Summarization",
    author = "Cheang, Chi  and
      Chan, Hou  and
      Wong, Derek  and
      Liu, Xuebo  and
      Li, Zhaocong  and
      Sun, Yanming  and
      Liu, Shudong  and
      Chao, Lidia",
    editor = "Bouamor, Houda  and
      Pino, Juan  and
      Bali, Kalika",
    booktitle = "Proceedings of the 2023 Conference on Empirical Methods in Natural Language Processing",
    month = dec,
    year = "2023",
    address = "Singapore",
    publisher = "Association for Computational Linguistics",
    url = "https://aclanthology.org/2023.emnlp-main.1007",
    doi = "10.18653/v1/2023.emnlp-main.1007",
    pages = "16205--16217",
}

@inproceedings{xu2024knowledge,
    title = "Knowledge Conflicts for {LLM}s: A Survey",
    author = "Xu, Rongwu  and
      Qi, Zehan  and
      Guo, Zhijiang  and
      Wang, Cunxiang  and
      Wang, Hongru  and
      Zhang, Yue  and
      Xu, Wei",
    editor = "Al-Onaizan, Yaser  and
      Bansal, Mohit  and
      Chen, Yun-Nung",
    booktitle = "Proceedings of the 2024 Conference on Empirical Methods in Natural Language Processing",
    month = nov,
    year = "2024",
    address = "Miami, Florida, USA",
    publisher = "Association for Computational Linguistics",
    url = "https://aclanthology.org/2024.emnlp-main.486",
    doi = "10.18653/v1/2024.emnlp-main.486",
    pages = "8541--8565",
}

@inproceedings{
wang2024resolving,
title={Resolving Knowledge Conflicts in Large Language Models},
author={Yike Wang and Shangbin Feng and Heng Wang and Weijia Shi and Vidhisha Balachandran and Tianxing He and Yulia Tsvetkov},
booktitle={First Conference on Language Modeling},
year={2024},
url={https://openreview.net/forum?id=ptvV5HGTNN}
}

@article{trivedi2022musique,
    title = "♫ {M}u{S}i{Q}ue: Multihop Questions via Single-hop Question Composition",
    author = "Trivedi, Harsh  and
      Balasubramanian, Niranjan  and
      Khot, Tushar  and
      Sabharwal, Ashish",
    editor = "Roark, Brian  and
      Nenkova, Ani",
    journal = "Transactions of the Association for Computational Linguistics",
    volume = "10",
    year = "2022",
    address = "Cambridge, MA",
    publisher = "MIT Press",
    url = "https://aclanthology.org/2022.tacl-1.31",
    doi = "10.1162/tacl_a_00475",
    pages = "539--554",
}

@inproceedings{NEURIPS2022_6f1d43d5,
 author = {Meng, Kevin and Bau, David and Andonian, Alex and Belinkov, Yonatan},
 booktitle = {Advances in Neural Information Processing Systems},
 editor = {S. Koyejo and S. Mohamed and A. Agarwal and D. Belgrave and K. Cho and A. Oh},
 pages = {17359--17372},
 publisher = {Curran Associates, Inc.},
 title = {Locating and Editing Factual Associations in GPT},
 url = {https://proceedings.neurips.cc/paper_files/paper/2022/file/6f1d43d5a82a37e89b0665b33bf3a182-Paper-Conference.pdf},
 volume = {35},
 year = {2022}
}

@article{wu2024faithful,
  title={How faithful are RAG models? Quantifying the tug-of-war between RAG and LLMs' internal prior},
  author={Wu, Kevin and Wu, Eric and Zou, James},
  journal={arXiv e-prints},
  pages={arXiv--2404},
  year={2024}
}

@article{kadavath2022language,
  title={Language models (mostly) know what they know},
  author={Kadavath, Saurav and Conerly, Tom and Askell, Amanda and Henighan, Tom and Drain, Dawn and Perez, Ethan and Schiefer, Nicholas and Hatfield-Dodds, Zac and DasSarma, Nova and Tran-Johnson, Eli and others},
  journal={arXiv preprint arXiv:2207.05221},
  year={2022}
}

@inproceedings{zhu2023calibration,
    title = "On the Calibration of Large Language Models and Alignment",
    author = "Zhu, Chiwei  and
      Xu, Benfeng  and
      Wang, Quan  and
      Zhang, Yongdong  and
      Mao, Zhendong",
    editor = "Bouamor, Houda  and
      Pino, Juan  and
      Bali, Kalika",
    booktitle = "Findings of the Association for Computational Linguistics: EMNLP 2023",
    month = dec,
    year = "2023",
    address = "Singapore",
    publisher = "Association for Computational Linguistics",
    url = "https://aclanthology.org/2023.findings-emnlp.654",
    doi = "10.18653/v1/2023.findings-emnlp.654",
    pages = "9778--9795",
}

@article{zhang2024calibrating,
  title={Calibrating the Confidence of Large Language Models by Eliciting Fidelity},
  author={Zhang, Mozhi and Huang, Mianqiu and Shi, Rundong and Guo, Linsen and Peng, Chong and Yan, Peng and Zhou, Yaqian and Qiu, Xipeng},
  journal={arXiv preprint arXiv:2404.02655},
  year={2024}
}

@InProceedings{pmlr-v139-zhao21c,
  title = 	 {Calibrate Before Use: Improving Few-shot Performance of Language Models},
  author =       {Zhao, Zihao and Wallace, Eric and Feng, Shi and Klein, Dan and Singh, Sameer},
  booktitle = 	 {Proceedings of the 38th International Conference on Machine Learning},
  pages = 	 {12697--12706},
  year = 	 {2021},
  editor = 	 {Meila, Marina and Zhang, Tong},
  volume = 	 {139},
  series = 	 {Proceedings of Machine Learning Research},
  month = 	 {18--24 Jul},
  publisher =    {PMLR},
  url = 	 {https://proceedings.mlr.press/v139/zhao21c.html}
}

@inproceedings{
zheng2024large,
title={Large Language Models Are Not Robust Multiple Choice Selectors},
author={Chujie Zheng and Hao Zhou and Fandong Meng and Jie Zhou and Minlie Huang},
booktitle={The Twelfth International Conference on Learning Representations},
year={2024},
url={https://openreview.net/forum?id=shr9PXz7T0}
}

@inproceedings{
turpin2023language,
title={Language Models Don't Always Say What They Think: Unfaithful Explanations in Chain-of-Thought Prompting},
author={Miles Turpin and Julian Michael and Ethan Perez and Samuel R. Bowman},
booktitle={Thirty-seventh Conference on Neural Information Processing Systems},
year={2023},
url={https://openreview.net/forum?id=bzs4uPLXvi}
}

@article{geva2021did,
  title={Did aristotle use a laptop? a question answering benchmark with implicit reasoning strategies},
  author={Geva, Mor and Khashabi, Daniel and Segal, Elad and Khot, Tushar and Roth, Dan and Berant, Jonathan},
  journal={Transactions of the Association for Computational Linguistics},
  volume={9},
  pages={346--361},
  year={2021},
  publisher={MIT Press One Rogers Street, Cambridge, MA 02142-1209, USA journals-info~…}
}

@inproceedings{clark2019boolq,
    title = "{B}ool{Q}: Exploring the Surprising Difficulty of Natural Yes/No Questions",
    author = "Clark, Christopher  and
      Lee, Kenton  and
      Chang, Ming-Wei  and
      Kwiatkowski, Tom  and
      Collins, Michael  and
      Toutanova, Kristina",
    editor = "Burstein, Jill  and
      Doran, Christy  and
      Solorio, Thamar",
    booktitle = "Proceedings of the 2019 Conference of the North {A}merican Chapter of the Association for Computational Linguistics: Human Language Technologies, Volume 1 (Long and Short Papers)",
    month = jun,
    year = "2019",
    address = "Minneapolis, Minnesota",
    publisher = "Association for Computational Linguistics",
    url = "https://aclanthology.org/N19-1300",
    doi = "10.18653/v1/N19-1300",
    pages = "2924--2936",
}

@inproceedings{jin2019pubmedqa,
    title = "{P}ub{M}ed{QA}: A Dataset for Biomedical Research Question Answering",
    author = "Jin, Qiao  and
      Dhingra, Bhuwan  and
      Liu, Zhengping  and
      Cohen, William  and
      Lu, Xinghua",
    editor = "Inui, Kentaro  and
      Jiang, Jing  and
      Ng, Vincent  and
      Wan, Xiaojun",
    booktitle = "Proceedings of the 2019 Conference on Empirical Methods in Natural Language Processing and the 9th International Joint Conference on Natural Language Processing (EMNLP-IJCNLP)",
    month = nov,
    year = "2019",
    address = "Hong Kong, China",
    publisher = "Association for Computational Linguistics",
    url = "https://aclanthology.org/D19-1259",
    doi = "10.18653/v1/D19-1259",
    pages = "2567--2577",
}

@inproceedings{yang2018hotpotqa,
    title = "{H}otpot{QA}: A Dataset for Diverse, Explainable Multi-hop Question Answering",
    author = "Yang, Zhilin  and
      Qi, Peng  and
      Zhang, Saizheng  and
      Bengio, Yoshua  and
      Cohen, William  and
      Salakhutdinov, Ruslan  and
      Manning, Christopher D.",
    editor = "Riloff, Ellen  and
      Chiang, David  and
      Hockenmaier, Julia  and
      Tsujii, Jun{'}ichi",
    booktitle = "Proceedings of the 2018 Conference on Empirical Methods in Natural Language Processing",
    month = oct # "-" # nov,
    year = "2018",
    address = "Brussels, Belgium",
    publisher = "Association for Computational Linguistics",
    url = "https://aclanthology.org/D18-1259",
    doi = "10.18653/v1/D18-1259",
    pages = "2369--2380",
}

@inproceedings{ho2020constructing,
    title = "Constructing A Multi-hop {QA} Dataset for Comprehensive Evaluation of Reasoning Steps",
    author = "Ho, Xanh  and
      Duong Nguyen, Anh-Khoa  and
      Sugawara, Saku  and
      Aizawa, Akiko",
    editor = "Scott, Donia  and
      Bel, Nuria  and
      Zong, Chengqing",
    booktitle = "Proceedings of the 28th International Conference on Computational Linguistics",
    month = dec,
    year = "2020",
    address = "Barcelona, Spain (Online)",
    publisher = "International Committee on Computational Linguistics",
    url = "https://aclanthology.org/2020.coling-main.580",
    doi = "10.18653/v1/2020.coling-main.580",
    pages = "6609--6625",
}

@inproceedings{joshi2017triviaqa,
    title = "{T}rivia{QA}: A Large Scale Distantly Supervised Challenge Dataset for Reading Comprehension",
    author = "Joshi, Mandar  and
      Choi, Eunsol  and
      Weld, Daniel  and
      Zettlemoyer, Luke",
    editor = "Barzilay, Regina  and
      Kan, Min-Yen",
    booktitle = "Proceedings of the 55th Annual Meeting of the Association for Computational Linguistics (Volume 1: Long Papers)",
    month = jul,
    year = "2017",
    address = "Vancouver, Canada",
    publisher = "Association for Computational Linguistics",
    url = "https://aclanthology.org/P17-1147",
    doi = "10.18653/v1/P17-1147",
    pages = "1601--1611",
}

@article{song2024large,
  title={Large Language Models are Skeptics: False Negative Problem of Input-conflicting Hallucination},
  author={Song, Jongyoon and Yu, Sangwon and Yoon, Sungroh},
  journal={arXiv preprint arXiv:2406.13929},
  year={2024}
}

@article{liu2024lost,
    title = "Lost in the Middle: How Language Models Use Long Contexts",
    author = "Liu, Nelson F.  and
      Lin, Kevin  and
      Hewitt, John  and
      Paranjape, Ashwin  and
      Bevilacqua, Michele  and
      Petroni, Fabio  and
      Liang, Percy",
    journal = "Transactions of the Association for Computational Linguistics",
    volume = "12",
    year = "2024",
    address = "Cambridge, MA",
    publisher = "MIT Press",
    url = "https://aclanthology.org/2024.tacl-1.9",
    doi = "10.1162/tacl_a_00638",
    pages = "157--173",
}

@inproceedings{wolf2020transformers,
    title = "Transformers: State-of-the-Art Natural Language Processing",
    author = "Thomas Wolf and Lysandre Debut and Victor Sanh and Julien Chaumond and Clement Delangue and Anthony Moi and Pierric Cistac and Tim Rault and Rémi Louf and Morgan Funtowicz and Joe Davison and Sam Shleifer and Patrick von Platen and Clara Ma and Yacine Jernite and Julien Plu and Canwen Xu and Teven Le Scao and Sylvain Gugger and Mariama Drame and Quentin Lhoest and Alexander M. Rush",
    booktitle = "Proceedings of the 2020 Conference on Empirical Methods in Natural Language Processing: System Demonstrations",
    month = oct,
    year = "2020",
    address = "Online",
    publisher = "Association for Computational Linguistics",
    url = "https://www.aclweb.org/anthology/2020.emnlp-demos.6",
    pages = "38--45"
}

@article{yu2024unleashing,
  title={Unleashing Multi-Hop Reasoning Potential in Large Language Models through Repetition of Misordered Context},
  author={Yu, Sangwon and Kim, Ik-hwan and Song, Jongyoon and Lee, Saehyung and Park, Junsung and Yoon, Sungroh},
  journal={arXiv preprint arXiv:2410.07103},
  year={2024}
}

@misc{Song2024ICiD,
publisher = {College of Seoul National University},
year = {2024},
title = {Input-output Consistency in Deep Learning based Conditional Text Generation},
language = {eng},
address = {Ph.D. dissertation, Dept. of ECE, Seoul National Univ.},
author = {Jongyoon Song},
}

@misc{chuang2024doladecodingcontrastinglayers,
      title={DoLa: Decoding by Contrasting Layers Improves Factuality in Large Language Models}, 
      author={Yung-Sung Chuang and Yujia Xie and Hongyin Luo and Yoon Kim and James Glass and Pengcheng He},
      year={2024},
      eprint={2309.03883},
      archivePrefix={arXiv},
      primaryClass={cs.CL},
      url={https://arxiv.org/abs/2309.03883}, 
}

@inproceedings{
chen2024inside,
title={{INSIDE}: {LLM}s' Internal States Retain the Power of Hallucination Detection},
author={Chao Chen and Kai Liu and Ze Chen and Yi Gu and Yue Wu and Mingyuan Tao and Zhihang Fu and Jieping Ye},
booktitle={The Twelfth International Conference on Learning Representations},
year={2024},
url={https://openreview.net/forum?id=Zj12nzlQbz}
}

@misc{simhi2025distinguishingignoranceerrorllm,
      title={Distinguishing Ignorance from Error in LLM Hallucinations}, 
      author={Adi Simhi and Jonathan Herzig and Idan Szpektor and Yonatan Belinkov},
      year={2025},
      eprint={2410.22071},
      archivePrefix={arXiv},
      primaryClass={cs.CL},
      url={https://arxiv.org/abs/2410.22071}, 
}

\clearpage
\appendices
\section{Details of Dataset}\label{app:details_of_dataset}
\begin{table}[t!]
    \centering
    \caption{Statistics of the evaluation sets categorized based on the state of parametric knowledge.}
    {\resizebox{0.92\columnwidth}{!}{
    \renewcommand{\arraystretch}{1.1}
    \begin{tabular}{l|ccc}
    \toprule       
       \textbf{Llama}   & \textbf{Parametric} & \textbf{Counter.} & \textbf{Absent}\\ \hline
       StrategyQA & 319 / 544 & 176 / 52 &  14 / 18 \\
       HotpotQA & 375 / 374 & 450 / 451 &  145 / 146 \\
       BoolQ & 507 / 346 & 167 / 65 &  50 / 40 \\
       MuSiQue & 193 / 189 & 667 / 668 &  331 / 329 \\
       2WikiMultiHopQA & 230 / 231 & 310 / 304 &  348 / 343 \\
       PubMedQA & 256 / 70 & 72 / 162 &  50 / 50 \\
       TriviaQA & 376 / 377 & 101 / 101 &  48 / 50 \\
    \hline
       \textbf{Qwen}   & \textbf{Parametric} & \textbf{Counter.} & \textbf{Absent}\\ \hline
       StrategyQA & 302 / 817 & 241 / 46 &  27 / 25 \\
       HotpotQA & 219 / 213 & 313 / 312 &  424 / 429 \\
       BoolQ & 429 / 471 & 307 / 52 &  50 / 46 \\
       MuSiQue & 88 / 89 & 304 / 308 &  785 / 794 \\
       2WikiMultiHopQA & 167 / 169 & 240 / 245 &  444 / 446 \\
       PubMedQA & 238 / 63 & 70 / 183 &  50 / 50 \\
       TriviaQA & 258 / 260 & 134 / 133 &  101 / 102 \\
    \hline
       \textbf{Mistral}   & \textbf{Parametric} & \textbf{Counter.} & \textbf{Absent}\\ \hline
       StrategyQA & 198 / 615 & 248 / 34 &  74 / 58 \\
       HotpotQA & 284 / 285 & 392 / 390 &  291 / 294 \\
       BoolQ & 352 / 401 & 261 / 50 &  58 / 50 \\
       MuSiQue & 98 / 94 & 467 / 475 &  604 / 596 \\
       2WikiMultiHopQA & 156 / 164 & 360 / 372 &  371 / 378 \\
       PubMedQA & 165 / 50 & 50 / 119 &  89 / 78 \\
       TriviaQA & 313 / 316 & 135 / 134 &  50 / 50 \\
    \hline
       \textbf{GPT-4o}   & \textbf{Parametric} & \textbf{Counter.} & \textbf{Absent}\\ \hline
       StrategyQA & 638 / 950 & 144 / 55 &  82 / 56 \\
       HotpotQA & 622 / 615 & 143 / 144 &  222 / 221 \\
       BoolQ & 940 / 615 & 168 / 57 &  50 / 50 \\
       MuSiQue & 375 / 375 & 296 / 300 &  511 / 519 \\
       2WikiMultiHopQA & 556 / 546 & 124 / 127 &  259 / 262 \\
       PubMedQA & 248 / 75 & 56 / 169 &  94 / 86 \\
       TriviaQA & 466 / 469 & 49 / 50 &  49 / 49 \\
    \bottomrule
    \end{tabular}
    }}    
    \label{tab:statistics}
\end{table}
We exclude samples where the number of tokens in the context exceeded 2,048 for all cases.
For StrategyQA we utilize the entire training set.
For HotpotQA, we employ the development set of the fullwiki version.
For BoolQ and 2WikiMultiHopQA, we sample 2,000 examples from the validation set. 
For subsets with insufficient samples, we further utilize the training set.
For MuSiQue, we use the ``answerable'' validation set. 
of reasoning steps from the training set.
For PubMedQA, we utilize the training set of ``pqa\_labeled''. 
For subsets with insufficient sample sizes, we incorporate 9,000 samples from the training set of ``pqa\_artificial''. 
For the \textit{absent} subset, we include samples from the remaining training set that are classified as ``Unanswerable'' when prompted to the corresponding model without chain-of-thought.
For TriviaQA, we sample 1,000 examples from the validation set of the ``rc.wikipedia'' subset. 
For subsets with insufficient samples, we utilize the remaining training and validation sets.

\section{Case Study}\label{app:case_study}
\begin{table}
\centering
\caption{Examples of Prediction Shift in Prompting Scenarios.}
{\small
\renewcommand{\arraystretch}{1.2}
\begin{tabular}{|p{\linewidth}|}
\hline
\texttt{\textbf{Failure Case of CoT Prompting}} \\
\texttt{- GPT-4o / MuSiQue / Parametric (Positive)} \\
\texttt{\textbf{Question}: Did the author of Elizabeth and After attend the University of Toronto?} \\
\texttt{\textbf{Original Prediction}: Yes.} \\
\texttt{\textbf{Prediction with CoT}: To determine if the author of ``Elizabeth and After'' attended the University of Toronto, we need to identify the author and their educational background. The author of ``Elizabeth and After'' is Matt Cohen. ... However, there is no widely available information indicating that he attended the University of Toronto. Answer: No.} \\
\hline
\texttt{\textbf{Success Case of Context Prompting}} \\ 
\texttt{- Qwen / TriviaQA / Counter. (Positive)} \\
\texttt{\textbf{Question}: Did Jan Francis play Penny in Just Good Friends?} \\
\texttt{\textbf{Original Prediction}: No.} \\
\texttt{\textbf{Prediction with Context}: [Title: Just Good Friends] Just Good Friends was a British sitcom written by ... Answer: Yes.} \\
\hline
\texttt{\textbf{Prediction Shift Case of IDK Prompting}} \\ 
\texttt{- Llama / PubMedQA / Absent (Positive)} \\
\texttt{\textbf{Question}: Does thrombomodulin influence the Survival of Patients with Non-Metastatic Colorectal Cancer through Epithelial-To-Mesenchymal Transition (EMT)?} \\
\texttt{\textbf{Original Prediction}: No.} \\
\texttt{\textbf{Prediction with IDK Option}: Unanswerable.} \\
\hline
\end{tabular}
}
\label{tab:case_study}
\end{table}
\begin{table*}
\centering
\caption{$\Delta$ and Weighted F1 results across datasets.}\label{tab:default_result}
\resizebox{\linewidth}{!}{
\begin{tabular}{l|c|cc|cc|cc|cc|cc|cc}
\hline
\toprule
\multicolumn{1}{c}{ } &
\multicolumn{1}{c}{ } &
\multicolumn{4}{c}{Parametric} &
\multicolumn{4}{c}{Counter-parametric} & 
\multicolumn{4}{c}{Absent} \\
\multicolumn{1}{c}{ } &
\multicolumn{1}{c}{ } &
\multicolumn{2}{c}{$\Delta$} &
\multicolumn{2}{c}{W.F1} &
\multicolumn{2}{c}{$\Delta$} &
\multicolumn{2}{c}{W.F1} &
\multicolumn{2}{c}{$\Delta$} &
\multicolumn{2}{c}{W.F1} \\
Model & Dataset & MCQA & YNQA & MCQA & YNQA & MCQA & YNQA & MCQA & YNQA & MCQA & YNQA & MCQA & YNQA \\
\hline
\multirow{7}{*}{Llama} & StrategyQA & 0.126 & 0.160 & \textbf{0.919} & 0.864 & 0.040 & 0.337 & 0.191 & \textbf{0.267} & 0.381 & 0.762 & \textbf{0.656} & 0.415 \\
& HotpotQA & $-$0.455 & 0.013 & \textbf{0.771} & 0.702 & $-$0.467 & 0.241 & \textbf{0.606} & 0.581 & $-$0.436 & 0.607 & \textbf{0.607} & 0.546 \\
& BoolQ & 0.012 & 0.188 & \textbf{0.940} & 0.829 & 0.056 & 0.281 & 0.213 & \textbf{0.225} & $-$0.125 & 0.795 & \textbf{0.638} & 0.332 \\
& MuSiQue & $-$0.404 & 0.396 & \textbf{0.742} & 0.662 & $-$0.261 & 0.609 & \textbf{0.558} & 0.504 & $-$0.204 & 0.819 & \textbf{0.560} & 0.467 \\
& 2WikiMultiHopQA & $-$0.171 & $-$0.124 & \textbf{0.670} & 0.572 & $-$0.307 & 0.522 & \textbf{0.574} & 0.520 & $-$0.170 & 0.551 & \textbf{0.606} & 0.463 \\
& PubMedQA & 0.038 & $-$0.087 & \textbf{0.965} & 0.872 & $-$0.007 & $-$0.094 & 0.073 & \textbf{0.298} & 0.040 & $-$0.240 & \textbf{0.864} & 0.716 \\
& TriviaQA & $-$0.438 & 0.147 & \textbf{0.847} & 0.828 & $-$0.535 & 0.258 & 0.639 & \textbf{0.708} & $-$0.567 & 0.525 & 0.536 & \textbf{0.616} \\
\hline
\multirow{7}{*}{Qwen} &StrategyQA & 0.193 & 0.358 & \textbf{0.967} & 0.882 & 0.123 & 0.482 & 0.054 & \textbf{0.148} & 0.492 & 0.886 & \textbf{0.519} & 0.381 \\
& HotpotQA & 0.156 & 0.310 & \textbf{0.815} & 0.675 & 0.132 & 0.658 & \textbf{0.651} & 0.494 & 0.199 & 0.840 & \textbf{0.635} & 0.424 \\
& BoolQ & 0.148 & 0.232 & \textbf{0.965} & 0.851 & 0.112 & 0.399 & 0.044 & \textbf{0.167} & 0.405 & 0.817 & \textbf{0.471} & 0.441 \\
& MuSiQue & 0.243 & 0.739 & \textbf{0.764} & 0.548 & 0.472 & 0.925 & \textbf{0.512} & 0.391 & 0.453 & 0.971 & \textbf{0.485} & 0.350 \\
& 2WikiMultiHopQA & 0.198 & 0.512 & \textbf{0.647} & 0.487 & 0.467 & 0.847 & \textbf{0.521} & 0.419 & 0.049 & 0.879 & \textbf{0.500} & 0.398 \\
& PubMedQA & $-$0.072 & 0.127 & \textbf{0.989} & 0.835 & $-$0.032 & 0.169 & 0.019 & \textbf{0.231} & 0.080 & 0.740 & \textbf{0.722} & 0.432 \\
& TriviaQA & $-$0.064 & 0.067 & \textbf{0.913} & 0.816 & 0.019 & 0.469 & \textbf{0.775} & 0.632 & 0.173 & 0.674 & \textbf{0.641} & 0.566 \\
\hline
\multirow{7}{*}{Mistral} & StrategyQA & 0.065 & $-$0.089 & \textbf{0.985} & 0.916 & 0.021 & $-$0.138 & 0.016 & \textbf{0.299} & 0.340 & 0.076 & 0.450 & \textbf{0.570} \\
& HotpotQA & $-$0.048 & $-$0.248 & \textbf{0.789} & 0.670 & 0.010 & $-$0.017 & \textbf{0.659} & 0.616 & 0.061 & 0.227 & \textbf{0.630} & 0.576 \\
& BoolQ & 0.080 & $-$0.065 & \textbf{0.992} & 0.883 & 0.017 & $-$0.164 & 0.041 & \textbf{0.336} & 0.276 & $-$0.026 & \textbf{0.658} & 0.580 \\
& MuSiQue & 0.023 & 0.225 & \textbf{0.745} & 0.647 & 0.229 & 0.297 & \textbf{0.566} & 0.536 & 0.387 & 0.554 & \textbf{0.513} & 0.505 \\
& 2WikiMultiHopQA & 0.468 & $-$0.477 & \textbf{0.542} & 0.517 & 0.248 & $-$0.009 & \textbf{0.519} & 0.518 & 0.431 & $-$0.072 & 0.516 & \textbf{0.525} \\
& PubMedQA & $-$0.184 & $-$0.242 & \textbf{0.995} & 0.923 & $-$0.020 & $-$0.383 & 0.003 & \textbf{0.082} & $-$0.224 & $-$0.646 & \textbf{0.546} & 0.453 \\
& TriviaQA & $-$0.401 & $-$0.151 & \textbf{0.858} & 0.827 & $-$0.272 & $-$0.010 & \textbf{0.700} & 0.647 & $-$0.360 & 0.280 & \textbf{0.735} & 0.633 \\
\hline
\multirow{7}{*}{GPT-4o} &StrategyQA & 0.004 & 0.078 & \textbf{0.976} & 0.925 & $-$0.128 & 0.072 & 0.161 & \textbf{0.181} & $-$0.083 & 0.516 & \textbf{0.601} & 0.529 \\
& HotpotQA & $-$0.208 & $-$0.036 & \textbf{0.897} & 0.828 & $-$0.157 & 0.002 & \textbf{0.696} & 0.638 & $-$0.156 & 0.243 & \textbf{0.731} & 0.611 \\
& BoolQ & $-$0.020 & 0.026 & \textbf{0.993} & 0.942 & $-$0.018 & 0.051 & 0.044 & \textbf{0.136} & $-$0.200 & 0.160 & \textbf{0.604} & 0.537 \\
& MuSiQue & $-$0.131 & 0.058 & \textbf{0.842} & 0.776 & 0.135 & 0.301 & \textbf{0.644} & 0.581 & $-$0.033 & 0.445 & \textbf{0.677} & 0.571 \\
& 2WikiMultiHopQA & $-$0.095 & $-$0.058 & \textbf{0.774} & 0.665 & $-$0.227 & $-$0.030 & \textbf{0.627} & 0.582 & 0.037 & 0.370 & \textbf{0.635} & 0.578 \\
& PubMedQA & $-$0.112 & $-$0.060 & \textbf{0.987} & 0.966 & $-$0.143 & $-$0.143 & 0.017 & \textbf{0.085} & $-$0.533 & $-$0.226 & \textbf{0.600} & 0.578 \\
& TriviaQA & $-$0.143 & 0.009 & \textbf{0.967} & 0.942 & $-$0.176 & 0.207 & \textbf{0.856} & 0.734 & $-$0.122 & 0.388 & \textbf{0.872} & 0.671 \\
\bottomrule
\end{tabular}}
\end{table*}
\begin{table*}
\centering
\caption{$\Delta$ and weighted F1 score Results across prompting scenarios}\label{tab:prompting_scenario_result}
\resizebox{\linewidth}{!}{
\begin{tabular}{l|c|cc|cc|cc|cc|cc|cc}
\hline
\toprule
\multicolumn{1}{c}{ } &
\multicolumn{1}{c}{ } &
\multicolumn{4}{c}{Parametric} &
\multicolumn{4}{c}{Counter-parametric} & 
\multicolumn{4}{c}{Absent} \\
\multicolumn{1}{c}{ } &
\multicolumn{1}{c}{ } &
\multicolumn{2}{c}{$\Delta$} &
\multicolumn{2}{c}{W.F1} &
\multicolumn{2}{c}{$\Delta$} &
\multicolumn{2}{c}{W.F1} &
\multicolumn{2}{c}{$\Delta$} &
\multicolumn{2}{c}{W.F1} \\
Model & Dataset & MCQA & YNQA & MCQA & YNQA & MCQA & YNQA & MCQA & YNQA & MCQA & YNQA & MCQA & YNQA \\
\hline
\multicolumn{14}{c}{\textit{Without Context}} \\
\hline
\multirow{4}{*}{Llama} & None & $-$0.185 & 0.099 & \textbf{0.836} & 0.761 & $-$0.212 & 0.308 & 0.408 & \textbf{0.443} & $-$0.154 & 0.638 & \textbf{0.546} & 0.508  \\
& +IDK & $-$0.231 & $-$0.093 & \textbf{0.780} & 0.773 & $-$0.212 & 0.026 & 0.376 & \textbf{0.476} & $-$0.200 & 0.053 & 0.549 & \textbf{0.572}  \\
& +CoT & $-$0.206 & 0.143 & \textbf{0.899} & 0.769 & $-$0.160 & 0.241 & 0.395 & \textbf{0.452} & $-$0.218 & 0.541 & \textbf{0.600} & 0.507  \\
& +IDK+CoT & $-$0.220 & 0.067 & \textbf{0.907} & 0.763 & $-$0.258 & 0.150 & 0.374 & \textbf{0.466} & $-$0.247 & 0.160 & \textbf{0.628} & 0.569  \\
\hline
\multirow{4}{*}{Qwen} & None & 0.115 & 0.335 & \textbf{0.866} & 0.728 & 0.185 & 0.564 & \textbf{0.368} & 0.355 & 0.264 & 0.830 & \textbf{0.568} & 0.427  \\
& +IDK & $-$0.093 & $-$0.249 & \textbf{0.872} & 0.808 & $-$0.029 & $-$0.044 & 0.372 & \textbf{0.474} & $-$0.021 & $-$0.034 & 0.514 & \textbf{0.727}  \\
& +CoT & $-$0.059 & 0.303 & \textbf{0.918} & 0.741 & $-$0.033 & 0.530 & \textbf{0.401} & 0.349 & $-$0.030 & 0.793 & \textbf{0.606} & 0.436  \\
& +IDK+CoT & $-$0.247 & $-$0.370 & \textbf{0.919} & 0.801 & $-$0.172 & $-$0.057 & 0.370 & \textbf{0.490} & $-$0.135 & $-$0.047 & 0.607 & \textbf{0.688}  \\
\hline
\multirow{4}{*}{Mistral} & None & 0.000 & $-$0.150 & \textbf{0.844} & 0.769 & 0.033 & $-$0.061 & 0.358 & 0.433 & 0.130 & 0.055 & \textbf{0.578} & 0.549  \\
& +IDK & $-$0.285 & $-$0.256 & \textbf{0.866} & 0.765 & $-$0.158 & $-$0.156 & 0.358 & \textbf{0.455} & $-$0.094 & $-$0.133 & \textbf{0.565} & 0.557  \\
& +CoT & 0.022 & 0.005 & \textbf{0.909} & 0.780 & 0.033 & 0.135 & 0.361 & \textbf{0.412} & 0.174 & 0.332 & \textbf{0.578} & 0.539  \\
& +IDK+CoT & $-$0.139 & $-$0.045 & \textbf{0.907} & 0.780 & $-$0.090 & 0.071 & \textbf{0.446} & 0.426 & $-$0.094 & 0.139 & \textbf{0.620} & 0.557  \\
\hline
\multirow{4}{*}{GPT-4o} & None & $-$0.101 & 0.002 & \textbf{0.919} & 0.863 & $-$0.102 & 0.066 & \textbf{0.435} & 0.420 & $-$0.156 & 0.271 & \textbf{0.674} & 0.582  \\
& +IDK & $-$0.166 & $-$0.128 & \textbf{0.925} & 0.866 & $-$0.120 & $-$0.049 & \textbf{0.435} & 0.433 & $-$0.103 & $-$0.002 & \textbf{0.679} & 0.627  \\
& +CoT & $-$0.056 & 0.048 & \textbf{0.937}& 0.907 & $-$0.008 & 0.179 & \textbf{0.442} & 0.410 & $-$0.052 & 0.419 & \textbf{0.679} & 0.578  \\
& +IDK+CoT & $-$0.125 & $-$0.031 & \textbf{0.958} & 0.913 & $-$0.029 & 0.074 & \textbf{0.460} & 0.420 & $-$0.078 & 0.134 & \textbf{0.732} & 0.610  \\
\hline
\multicolumn{14}{c}{\textit{With Context}} \\
\hline
\multirow{4}{*}{Llama} & None & $-$0.054 & 0.040 & \textbf{0.939} & 0.879 & $-$0.026 & 0.182 & \textbf{0.772} & 0.746 & $-$0.019 & 0.166 & \textbf{0.885} & 0.849  \\
& +IDK & $-$0.088 & 0.021 & \textbf{0.917} & 0.881 & $-$0.072 & 0.145 & 0.736 & \textbf{0.754} & $-$0.063 & 0.066 & 0.865 & \textbf{0.870}  \\
& +CoT & $-$0.088 & 0.072 & \textbf{0.954} & 0.891 & $-$0.092 & 0.185 & \textbf{0.835} & 0.746 & $-$0.006 & 0.160 & \textbf{0.938} & 0.853  \\
& +IDK+CoT & $-$0.149 & 0.044 & \textbf{0.947} & 0.888 & $-$0.177 & 0.162 & \textbf{0.833} & 0.745 & $-$0.141 & 0.102 & \textbf{0.922} & 0.860  \\
\hline
\multirow{4}{*}{Qwen} & None & 0.148 & 0.194 & \textbf{0.911} & 0.850 & 0.199 & 0.346 & \textbf{0.726} & 0.686 & 0.210 & 0.411 & \textbf{0.832} & 0.725  \\
& +IDK & $-$0.021 & $-$0.255 & \textbf{0.913} & 0.886 & $-$0.037 & $-$0.118 & 0.723 & \textbf{0.746} & $-$0.067 & $-$0.165 & \textbf{0.850} & 0.806  \\
& +CoT & 0.033 & 0.122 & \textbf{0.956} & 0.864 & 0.078 & 0.284 & \textbf{0.815} & 0.707 & 0.098 & 0.357 & \textbf{0.902} & 0.735 \\
& +IDK+CoT & $-$0.145 & $-$0.413 & \textbf{0.995} & 0.860 & $-$0.180 & $-$0.270 & \textbf{0.831} & 0.742 & $-$0.151 & $-$0.299 & \textbf{0.916} & 0.784  \\
\hline
\multirow{4}{*}{Mistral} & None & 0.007 & $-$0.102 & \textbf{0.920} & 0.864 & $-$0.015 & $-$0.038 & \textbf{0.713} & 0.704 & 0.053 & 0.067 & \textbf{0.848} & 0.790  \\
& +IDK & $-$0.218 & $-$0.158 & \textbf{0.929} & 0.862 & $-$0.169 & $-$0.110 & \textbf{0.722} & 0.709 & $-$0.217 & $-$0.050 & \textbf{0.868} & 0.801  \\
& +CoT & $-$0.020 & $-$0.051 & \textbf{0.944} & 0.855 & 0.022 & 0.007 & \textbf{0.733} & 0.698 & 0.034 & 0.133 & \textbf{0.864} & 0.774  \\
& +IDK+CoT & $-$0.152 & $-$0.088 & \textbf{0.956} & 0.854 & $-$0.122 & $-$0.030 & \textbf{0.751} & 0.705 & $-$0.164 & 0.050 & \textbf{0.913} & 0.787  \\
\hline
\multirow{4}{*}{GPT-4o} & None & $-$0.036 & $-$0.003 & \textbf{0.976} & 0.950 & 0.005 & 0.060 & \textbf{0.759} & 0.742 & 0.006 & 0.150 & \textbf{0.897} & 0.836  \\
& +IDK & $-$0.137 & $-$0.002 & \textbf{0.981} & 0.884 & $-$0.117 & 0.049 & \textbf{0.768} & 0.744 & $-$0.161 & 0.075 & \textbf{0.940} & 0.871  \\
& +CoT & $-$0.026 & 0.022 & \textbf{0.981} & 0.946 & 0.047 & 0.152 & \textbf{0.759} & 0.716 & 0.024 & 0.181 & \textbf{0.896} & 0.828  \\
& +IDK+CoT & $-$0.101 & 0.018 & \textbf{0.984} & 0.956 & $-$0.042 & 0.101 & \textbf{0.773} & 0.739 & $-$0.089 & 0.103 & \textbf{0.949} & 0.867  \\
\bottomrule
\end{tabular}}
\end{table*}
Table \ref{tab:case_study} presents examples in which prompting scenarios change the prediction of LLMs. 
An interesting observation arises from samples where a negative response is generated during the CoT process. 
In probing parametric knowledge on MuSiQue through free-form question answering, we observed that the model initially produced the correct answer. 
However, during the reasoning process, the model appeared to reflect its confidence in the underlying knowledge, which in turn could shift the binary decision in the wrong direction.

\section{Raw Results of Experiments}\label{app:default_raw_results}

Table \ref{tab:default_result} presents the results obtained from the experiment in Section \ref{sec:initial_observation}, including the $\Delta$ values for MCQA and YNQA types, along with the weighted F1 scores.
Table \ref{tab:prompting_scenario_result} presents the results obtained from the experiment in Section \ref{sec:prompting_scenario}.
We define the cases where the model generates a positive or negative response for a positive sample as true positive and false negative, respectively. 
Similarly, when the model generates a positive or negative response for a negative sample, we refer to them as false positive and true negative, respectively.

\section{Prompts}
The prompts used for GPT-4o in the evaluation set construction pipeline are shown in Tables \ref{prompt:binary_decision_data_conversion}, \ref{prompt:parametric_knowledge_probing}, and \ref{prompt:others}.

\begin{table*}
\centering
\caption{GPT-4o prompt used in binary decision data conversion. Few-shot examples are sampled from 2WikiMultihopQA ~\cite{ho2020constructing} and MuSiQue ~\cite{trivedi2022musique}.}
{\small
\renewcommand{\arraystretch}{1.2}
\begin{tabular}{p{15cm}}
\toprule
\texttt{\textbf{Binary Decision Data Conversion}} \\
\texttt{<User>} \\
\texttt{Given a question, a correct answer, and a wrong answer, write a pair of questions where the answer is `Yes' (Yes-Question) and `No' (No-Question). Do not omit any information in the given question.} \\ \\
\texttt{[Examples (begin)]} \\
\texttt{[Input]} \\
\texttt{Question: Which country the director of film Hotel By The Hour is from?} \\
\texttt{Correct Answer: Austria} \\
\texttt{Wrong Answer: United States} \\
\texttt{[Output]} \\
\texttt{Yes-Question: Is the director of film Hotel By The Hour from Austria?} \\
\texttt{No-Question: Is the director of film Hotel By The Hour from United States?} \\ \\
\texttt{[Input]} \\
\texttt{Question: Which film has the director born later, Life Hits or It'S In The Air?} \\
\texttt{Correct Answer: Life Hits} \\
\texttt{Wrong Answer: It'S In The Air} \\
\texttt{[Output]} \\
\texttt{Yes-Question: Is the director of Life Hits born later than the director of It's In The Air?} \\
\texttt{No-Question: Is the director of It's In The Air born later than the director of Life Hits?} \\ \\
\texttt{[Input]} \\ 
\texttt{Question: A country's military branch, which in the US contains the Air Defense Artillery, was unprepared for the invasion of Hana Mandlikova's birth country. When was the word ``Slavs'' used in the national anthem of the unprepared country?} \\
\texttt{Correct Answer: 1943-1992} \\
\texttt{Wrong Answer: 1968-2003} \\
\texttt{[Output]} \\ 
\texttt{Yes-Question: A country's military branch, which in the US contains the Air Defense Artillery, was unprepared for the invasion of Hana Mandlikova's birth country. Was the word ``Slavs'' used in the national anthem of the unprepared country from 1943 to 1992?} \\
\texttt{No-Question: A country's military branch, which in the US contains the Air Defense Artillery, was unprepared for the invasion of Hana Mandlikova's birth country. Was the word ``Slavs'' used in the national anthem of the unprepared country from 1968 to 2003?} \\
\texttt{[Examples (end)]} \\ 
\texttt{[Input]} \\ 
\texttt{Question: \textit{\{question\}}} \\
\texttt{Correct Answer: \textit{\{correct answer\}}} \\
\texttt{Wrong Answer: \textit{\{wrong answer\}}} \\
\texttt{[Output]} \\
\bottomrule
\end{tabular}
}
\label{prompt:binary_decision_data_conversion}
\end{table*}
\begin{table*}
\centering
\caption{GPT-4o prompt used in parametric knowledge probing and YNQA evaluation.}
{\small
\renewcommand{\arraystretch}{1.2}
\begin{tabular}{p{15cm}}
\toprule
\texttt{\textbf{Parametric Knowledge Probing: Yes-no QA}} \\
\texttt{<User>} \\
\texttt{Your role is to select the correct statement among the two statements according to your knowledge. If you don't know which statement is correct, choose the option corresponding to ``I don't know''.}\\
\texttt{Please simply answer with ``(A)'', ``(B)'', or ``(C)''.} \\
\texttt{\textit{\{options\}}} \\
\texttt{<Assistant>} \\
\texttt{Let's think step by step. }\texttt{\{\textit{response}\}} \\
\texttt{<User>} \\
\texttt{Return only the answer with  ``(A)'', ``(B)'', or ``(C)'' after `Answer:'} \\
\texttt{<Assistant>} \\
\texttt{Answer: } \\
\hline
\texttt{\textbf{Parametric Knowledge Probing: Short-answer QA}} \\
\texttt{<User>} \\
\texttt{Answer the question. Write only the answer in a few words after `Answer:'.} \\
\texttt{If you cannot answer the question, please answer with ``Unanswerable''.}\\
\texttt{Question: \{\textit{question}\}} \\
\texttt{<Assistant>} \\
\texttt{Let's think step by step. }\texttt{\{\textit{response}\}} \\
\texttt{<User>} \\
\texttt{Return only the answer in a few words or ``Unanswerable'' after `Answer:'.} \\
\texttt{<Assistant>} \\
\texttt{Answer: } \\
\hline
\texttt{\textbf{YNQA Evaluation}} \\
\texttt{<User>} \\
\texttt{You are given a question and you MUST answer with Yes or No based on your knowledge (w/ context) and the given context. (w/ ``unanswerable'' option) If you don't know the answer, please respond with `Answer: Unanswerable'.} \\
\texttt{(w/ context) Context: \{\textit{context}\}} \\
\texttt{Question: \{\textit{question}\}} \\
\texttt{<Assistant>} \\
\texttt{(w/ chain-of-thoughts) Let's think step by step. }\texttt{\{\textit{response}\}} \\
\texttt{(w/o chain-of-thoughts) Answer:} \\
\texttt{<User>} \\
\texttt{(w/ chain-of-thoughts \& w/ ``unanswerable'' option) Return only the answer with Yes, No, or Unanswerable after `Answer:'.} \\
\texttt{(w/ chain-of-thoughts \& w/o ``unanswerable'' option) Return only the answer with Yes or No after `Answer:'.} \\
\texttt{<Assistant>} \\
\texttt{(w/ chain-of-thoughts) Answer: } \\
\bottomrule
\end{tabular}
}
\label{prompt:parametric_knowledge_probing}
\end{table*}
\begin{table*}
\centering
\caption{GPT-4o prompt used in statement \& negation conversion and wrong answer generation. Few-shot examples in the statement \& negation conversion are sampled from StrategyQA ~\cite{geva2021did}.}
{\small
\renewcommand{\arraystretch}{1.2}
\begin{tabular}{p{15cm}}
\toprule
\texttt{\textbf{Statement \& Negation Conversion}} \\
\texttt{<User>} \\
\texttt{Convert the given question into a statement and then rewrite the statement to express the exact opposite meaning. Do not omit any information in the given question.} \\ \\
\texttt{[Example (begin)]} \\
\texttt{Question: Would the top of Mount Fuji stick out of the Sea of Japan?} \\
\texttt{Statement: The top of Mount Fuji would stick out of the Sea of Japan.} \\
\texttt{Opposite: The top of Mount Fuji would sink in the Sea of Japan.} \\ \\
\texttt{Question: Is there a warthog on Broadway?} \\
\texttt{Statement: There is a warthog on Broadway.} \\
\texttt{Opposite: There is no warthog on Broadway.} \\ \\
\texttt{Question: Could someone with fine motor control issues benefit from an altered keyboard layout?} \\
\texttt{Statement: Someone with fine motor control issues could benefit from an altered keyboard layout.} \\
\texttt{Opposite: No one with fine motor control issues could benefit from an altered keyboard layout.} \\
\texttt{[Example (end)]} \\
\texttt{[Input]} \\ 
\texttt{Question: \{\textit{question}}\} \\
\hline
\texttt{\textbf{Wrong Answer Generation}} \\
\texttt{<User>} \\
\texttt{Using the Context, contaminate the Answer to be wrong for the given Question.} \\
\texttt{Question: \textit{\{question\}}} \\
\texttt{Context: \textit{\{context\}}} \\ 
\texttt{Answer: \textit{\{ground truth\}}} \\
\texttt{Contaminated answer:} \\
\bottomrule
\end{tabular}
}
\label{prompt:others}
\end{table*}

\vfill

\end{document}